\documentclass[Afour,sageh,times]{sagej}

\usepackage{array}
\usepackage{multirow}
\usepackage{xcolor}
\usepackage{siunitx}
\usepackage{url}
\usepackage[hidelinks]{hyperref}
\usepackage{cleveref}

\newcommand{\bench}{MVSim-LIO}
\graphicspath{{figs/}}
\makeatletter
\newenvironment{dataavail}%
{\subsection*{\normalsize\sagesf\bfseries Data availability statement}\begin{refsize}\noindent}%
{\end{refsize}}
\makeatother

\begin{document}

\title{\bench{}: A Simulated LiDAR-Inertial Odometry Benchmark\\with Exact Trajectory and Map Ground Truth}

\runninghead{G\'omez-Espinosa et al.}

\author{Ana G\'omez-Espinosa\affilnum{1}, Jos\'e Luis Torres-Moreno\affilnum{2} and
Jos\'e Luis Blanco-Claraco\affilnum{2}}

\affiliation{\affilnum{1}Department of Informatics, University of Almer\'ia, Spain\\
\affilnum{2}Department of Engineering and CIESOL, University of Almer\'ia, Spain}

\corrauth{Jos\'e Luis Blanco-Claraco, Department of Engineering, University of
Almer\'ia, 04120 Almer\'ia, Spain.}

\email{jlblanco@ual.es}

\begin{abstract}
Real-world benchmarks for LiDAR-inertial odometry (LIO) measure estimators
against a reference that is itself an estimate (GNSS/INS, motion capture or a
laser scan). At centimeter level, discrepancies cannot be confidently
attributed to the method, the reference or the sensor model. We present
\bench{}, a simulated LIO benchmark whose ground truth is exact by
construction: the trajectory is the simulator's input, and the map is the
geometry the sensors observe. Fifteen worlds, from streets and a tunnel to a
forest and multi-level buildings, are traversed by 30 six-degrees-of-freedom
trajectories (\SI{20.1}{\kilo\meter}) of five carriers (car, ground robot,
legged robot, quadruped and hand-held rig). A ray-traced spinning LiDAR with
per-point timestamps and semantic labels, and a motion-consistent
\SI{200}{\hertz} IMU, are rendered in paired variants sharing the motion and
random draws: global or rolling shutter, three LiDAR models, and three noise
levels, up to a realistic error model (range noise, beam footprint, mixed
returns, dropouts and per-ring calibration errors) with most parameter ranges from
real LiDARs. We evaluate eight LiDAR-only and LIO configurations with
trajectory and map metrics. Over Gaussian range noise, the realistic error
model raises the APE of three methods by about a third and of the others by
at most \SI{16}{\percent}, and changes the ranking little; de-skewing absorbs
most of the rolling shutter; and most failures occur in two degenerate
scenes, an open field and a tunnel. The data may also serve learning-based
methods needing exact ground truth with realistic sensor errors. Data and
tools are released at
\url{https://github.com/MOLAorg/mvsim_lio_benchmark}.
\end{abstract}


\maketitle

\section{Introduction}
\label{sec:intro}

LiDAR odometry (LO) and LiDAR-inertial odometry (LIO) have reached a maturity
where the leading methods often differ by a few centimeters over hundreds of
meters on public datasets \citep{xu2022fastlio2,chen2023dlio,koide2024glim,vizzo2023kissicp,blanco2025mola}.
At that level, the error of the reference trajectory is often comparable to the
differences between methods. The reference
trajectory of a real dataset comes from another sensor or estimator: RTK
GNSS fused with an INS \citep{geiger2012kitti,kim2020mulran}, a laser
tracker \citep{nguyen2022ntuviral}, motion capture or a total
station \citep{helmberger2022hilti}, or the registration of each scan against a
prior terrestrial laser scan (TLS) of the site \citep{ramezani2020newer,zhang2023hiltioxford,tao2025spires}.
Each source has an error budget of millimeters to centimeters, covers only part
of the sequence or part of the state (e.g., a position without orientation),
and must be calibrated and synchronized to the sensor rig. Errors in extrinsics,
time offsets or the reference itself end up in the score of every method, with
no way to separate them from the estimator's own error. Map ground truth is
scarcer still: it requires a survey-grade scan of the whole site, which few
datasets provide, and which is itself registered with finite accuracy.

In simulation, these error sources do not exist by definition.
If the sensor data is
rendered from a known trajectory in a known scene, the trajectory and the
scene geometry are known exactly, and every factor (sensor model, noise,
shutter, scene, motion) can be changed while everything else stays fixed.
One could object that simulated sensors differ from real ones; we address
this gap with an error model checked against five real LiDARs
(\cref{sec:validation}) and a comparison of rankings on real data
(\cref{sec:discussion}). LIO also
needs two properties that general-purpose LiDAR simulation does not
guarantee: the timing of each point within a rotating sweep, which simulators
approximate with discrete steps or sectors per
sweep \citep{deschaud2021kitticarla,han2023carlaloc},
and IMU readings that are consistent with the rendered motion.

The same data may also serve learning-based methods, for
odometry \citep{sun2023transfusionodom} or perception \citep{xiao2022synlidar},
which need precise ground truth for training and evaluation together with
realistic sensor errors in their inputs. Real datasets provide poses and labels
only up to the errors of their references and annotations, and noise-free
simulated data leave a gap to real sensors. In \bench{}, poses and per-point
labels are exact, mixed returns are labeled as outliers, and every OS1-32
sweep with the realistic error model has a noise-free counterpart rendered
from the same poses.

This paper presents \bench{}, a simulated LIO benchmark with exact ground
truth and scenes chosen to be hard for current methods. Its contributions
are:
\begin{itemize}
\item A dataset of 15 worlds and 30 trajectories of five carriers
(\SI{20.1}{\kilo\meter}, \SI{148}{\minute}) covering structured,
unstructured, repetitive, multi-level and geometrically degenerate scenes,
rendered by an offline analytic ray tracer as paired variants (noise level
$\times$ shutter model, and three LiDAR models) with per-point timestamps,
semantic and instance labels, and a \SI{200}{\hertz} IMU (\cref{sec:bench}).
\item Exact ground truth for the pose of every sensor frame at any time, and
reference maps rendered noise-free at the true poses, together with checks
that quantify how closely the sensor data agrees with them (\cref{sec:gt}).
\item A realistic LiDAR error model with range-dependent noise, a beam model
(footprint spread on oblique surfaces and thin objects, mixed returns at depth
discontinuities) and long-range dropouts, checked against real LiDAR models,
plus per-ring calibration errors of assumed magnitude
(\cref{sec:noise,sec:validation}).
\item A baseline evaluation of eight LO and LIO configurations with trajectory
and map metrics, paired comparisons of the noise level, the shutter model and
the sensor, a method-independent observability analysis of the degenerate
scenes, and a pairwise comparison of four methods with 24 real sequences
(\cref{sec:observability,sec:protocol,sec:results,sec:discussion}).
\item An open release of the worlds, trajectories, generator scripts, ground
truth tools, method configurations and evaluation code, so that any variant
can be regenerated deterministically or extended with new sensors and noise
levels.
\end{itemize}

\section{Related Work}
\label{sec:related}

\subsection{Real-world LiDAR-inertial datasets}
KITTI \citep{geiger2012kitti,geiger2013kitti} and MulRan \citep{kim2020mulran}
provide driving sequences with GNSS/INS references; KITTI's were not accurate
enough to aggregate its scans into consistent maps \citep{behley2019semantickitti},
and its odometry release has no per-point timestamps. Newer
College \citep{ramezani2020newer,zhang2021newercollege},
Hilti-Oxford \citep{zhang2023hiltioxford} and Oxford Spires \citep{tao2025spires}
register scans against a prior TLS map to obtain a high quality reference trajectory, and
are among the few to release the reference map; their reference accuracy is in
the millimeter to centimeter range, but only where the TLS covers the site.
Hilti-Oxford scores its challenge sequences only at a few millimeter-accurate
control points each, and the earlier Hilti challenge \citep{helmberger2022hilti}
at motion-capture or total-station positions. NTU VIRAL \citep{nguyen2022ntuviral},
M2DGR \citep{yin2022m2dgr}, BotanicGarden \citep{liu2024botanic},
VBR \citep{brizi2024vbr}, MCD \citep{nguyen2024mcd}, HeLiPR \citep{jung2024helipr},
FusionPortableV2 \citep{wei2025fusionportablev2}, the Hilti 2023
challenge \citep{nair2024hilti2023} and GEODE \citep{chen2025geode} broaden the
platforms, sensors and environments, the latter focusing on degenerate
scenes. In all of them the reference is an
estimate with its own errors and calibration, and the sensor noise, the scene
and the motion cannot be changed independently; refer to \cref{tab:datasets}.

\begin{table*}[t]
\centering
\caption{Real and simulated LiDAR(-inertial) datasets. Size: path length, or
sequences $\times$ duration where no length is given. Per-point time: the
firing time of each point as released. Variants: the same trajectory released
with different sensor or scene settings; a length $\times n$: the same paths
in $n$ variants or conditions.}
\label{tab:datasets}
\footnotesize
\setlength{\tabcolsep}{3pt}
\newcolumntype{P}[1]{>{\raggedright\arraybackslash}p{#1}}
\begin{tabular}{@{}P{2.4cm}P{1.25cm}lP{1.15cm}P{1.6cm}P{2.3cm}P{1.35cm}P{1.45cm}P{2.1cm}@{}}
\toprule
Dataset & Platform & Size & IMU & Per-point time & Reference trajectory & Reference map & Variants & Other data \\
\midrule
\multicolumn{9}{@{}l}{\emph{Real}} \\
KITTI \citep{geiger2012kitti} & car & \SI{39.2}{\kilo\meter} & raw data only & no & GNSS/INS & -- & -- & stereo cameras \\
MulRan \citep{kim2020mulran} & car & \SI{124}{\kilo\meter} & yes & no & GNSS/INS + SLAM & -- & -- & radar \\
Newer College \citep{ramezani2020newer} & hand-held & \SI{2.2}{\kilo\meter} & yes & yes & scans registered to TLS & TLS & -- & cameras \\
Hilti 2021 \citep{helmberger2022hilti} & hand-held & 12 seq. & yes & yes & mocap, total station (sparse) & -- & -- & cameras \\
Hilti-Oxford \citep{zhang2023hiltioxford} & hand-held & 16 seq. & yes & yes & TLS control points (sparse) & TLS & -- & cameras \\
Oxford Spires \citep{tao2025spires} & backpack & \SI{9.6}{\kilo\meter} & yes & yes & scans registered to TLS & TLS & -- & cameras \\
NTU VIRAL \citep{nguyen2022ntuviral} & aerial & \SI{1.9}{\kilo\meter} & yes & yes & laser tracker (position) & -- & -- & cameras, UWB \\
\midrule
\multicolumn{9}{@{}l}{\emph{Simulated}} \\
UTIAS \citep{yoon2019mapless} & car & 2 $\times$ \SI{5}{\minute} & -- & from sweep start/end & simulator, \SI{10}{\hertz} & -- & -- & traffic, labels \\
KITTI-CARLA \citep{deschaud2021kitticarla} & car & 7 $\times$ \SI{500}{\second} & -- & per \SI{1}{\milli\second} slice & simulator, \SI{1}{\kilo\hertz} & -- & -- & cameras, traffic, labels \\
CARLA-Loc \citep{han2023carlaloc} & car & \SI{6.4}{\kilo\meter} $\times$ 6 & with noise & no (10 sectors) & simulator, \SI{200}{\hertz} & -- & weather, traffic & cameras, radar, GNSS, labels \\
TartanGround \citep{patel2025tartanground} & wheeled, legged & 878 traj. & noise-free & no (from depth images) & simulator, \SI{10}{\hertz} & mesh samples & -- & cameras, labels, weather \\
\bench{} (ours) & 5 carriers & \SI{20.1}{\kilo\meter} $\times$ 9 & with noise & yes & simulator input, continuous & noise-free, \SI{10}{\centi\meter} voxels & noise, shutter, LiDAR & labels \\
\bottomrule
\end{tabular}
\end{table*}

\subsection{Simulated data for odometry and mapping}
General-purpose simulators such as Gazebo \citep{koenig2004gazebo},
CARLA \citep{dosovitskiy2017carla} and AirSim \citep{shah2017airsim} have
produced point cloud datasets for segmentation and
mapping \citep{deschaud2021pariscarla,xiao2022synlidar}; dedicated simulators
target point-level realism \citep{winiwarter2022helios,kong2023marsim}.
Simulated, motion-distorted sweeps with exact poses are not new: the UTIAS
set \citep{yoon2019mapless} fires each laser of a simulated HDL-64E once per
CARLA step, and KITTI-CARLA \citep{deschaud2021kitticarla} renders each sweep
in \SI{1}{\milli\second} slices; neither has an IMU or sensor noise.
CARLA-Loc \citep{han2023carlaloc} adds an IMU derived from the ego motion,
with white noise and bias random walk, and evaluates FAST-LIO2 among other
methods; it assembles each sweep from ten sectors and adds one level of
Gaussian range noise, and its twin sequences vary the weather and the traffic
along identical paths. Exact ground truth from simulation is also established
for RGB-D and visual SLAM \citep{handa2014iclnuim,wang2020tartanair}:
TartanAir \citep{wang2020tartanair} and TartanGround \citep{patel2025tartanground}
showed the value of hard, photo-realistic sequences, with LiDAR resampled from
rendered depth images (no sweep timing) and a noise-free IMU. Simulated LiDAR
with an IMU also appears in place recognition \citep{song2025cns}, with
deskewed sweeps and an ideal IMU, and in learned
odometry \citep{sun2023transfusionodom}. Learned LiDAR
simulation \citep{manivasagam2020lidarsim} reaches high realism for driving
scenes at the cost of a reconstruction of real sites. What \bench{} adds is
the combination of per-point timing of a rotating sensor, an IMU consistent
with the rendered motion, and paired variants that change one factor at a time:
the shutter, the LiDAR model or the noise level, whose \emph{noisy} step
changes the LiDAR and IMU noise together (\cref{tab:datasets}); in exchange, it has no cameras, weather or
dynamic objects, which CARLA-Loc and TartanGround provide.

\subsection{LiDAR error sources}
The precision of a range measurement degrades with range and with the
incidence angle, and surfaces seen at grazing angles return weaker
echoes \citep{soudarissanane2011scanning}; on an oblique surface, the beam
footprint also skews the return waveform and biases the
range \citep{laconte2019bias}. Where a beam straddles a depth discontinuity,
its return can fall between both surfaces (mixed
pixels) \citep{hebert1992mixed,tuley2005artifacts}. Multi-beam sensors carry per-ring
calibration errors in elevation, azimuth and range offset, which distort
each ring systematically \citep{glennie2010hdl64}. Our error model
reproduces these effects geometrically (\cref{sec:noise}).

\section{The Benchmark}
\label{sec:bench}

\begin{figure*}[t]
\centering
\includegraphics[width=\textwidth]{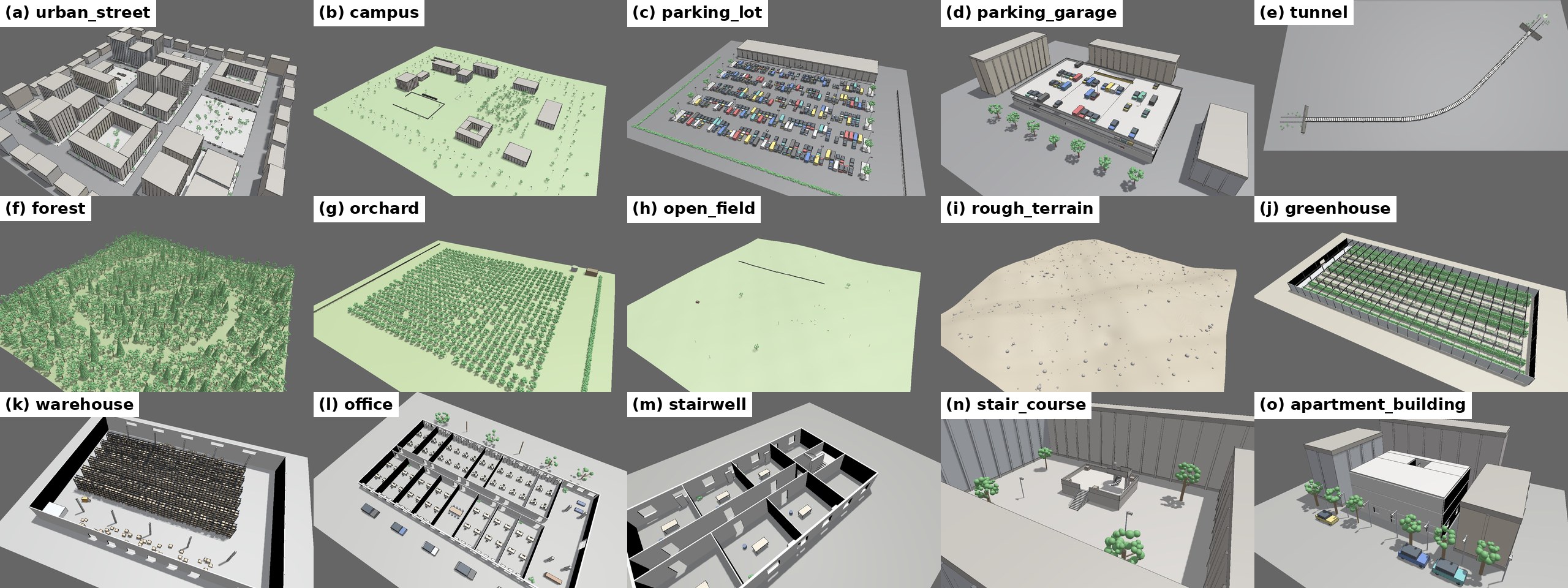}
\caption{The 15 worlds of \bench{}, rendered by MVSim from the same primitives
that the ray tracer intersects (indoor worlds shown as cutaways; the datasets include their roofs and ceilings).}
\label{fig:gallery}
\end{figure*}

\subsection{Simulation pipeline}
Worlds are described as MVSim \citep{blanco2023mvsim} world files generated by
Python scripts, which are the single source of truth for both the geometry and
the trajectories. For live simulation, MVSim renders its LiDARs on the GPU
from depth images, which is efficient for arbitrary 3D scenes and meshes, but
whose output may depend on the graphics hardware and driver. For this
benchmark, sensor data is instead rendered offline on the CPU by
\texttt{mvsim-dataset-gen}, which ray-traces exactly a simplified, analytic
geometry of the world: vertical prisms (walls, buildings, racks, steps), cylinders (trunks,
posts, columns), spheres (canopies built from clusters of small spheres),
wedges (ramps, roofs), horizontal planes and height fields (terrain). There
is no tessellation and no depth-buffer quantization as if rendered on a GPU: every range is the exact
distance to the analytic surface along the ray, up to machine
precision. Rendering is deterministic for a given world, trajectory, sensor
description and random seed.

\subsection{Worlds}
\Cref{fig:gallery} shows the 15 worlds and \cref{tab:worlds} summarizes what
each one stresses: structured, repetitive, multi-level and unstructured
scenes, and two geometrically degenerate
ones \citep{zhang2016degeneracy,tuna2024xicp}: a \SI{735}{\meter} road tunnel (the car drives through it, turns around outside
and drives back), where only sparse fixtures constrain
translation along the axis, and an open field where horizontal motion rests on
the gentle relief of the ground and a few distant objects.

\begin{table}[t]
\centering
\caption{Worlds and what they stress.}
\label{tab:worlds}
\footnotesize
\setlength{\tabcolsep}{3pt}
\begin{tabular}{@{}lp{0.70\columnwidth}@{}}
\toprule
World & Characteristics \\
\midrule
urban\_street & 3$\times$3 city blocks, facades, street furniture; well constrained, loops \\
campus & $500\times\SI{400}{\meter}$, buildings, park, sports field; long straights with sparse structure \\
parking\_lot & rows of similar cars; large open aisles \\
parking\_garage & three levels linked by \SI{16}{\percent} ramps; vertical drift \\
tunnel & \SI{735}{\meter} tunnel; axial degeneracy \\
forest & 1350 trees on hilly terrain; unstructured clutter, pitch and roll \\
orchard & 23 rows of fruit trees; aliasing across and along rows \\
open\_field & grassland with a few distant objects; horizontal degeneracy \\
rough\_terrain & steep hills, craters, boulders; fast pitch and roll \\
greenhouse & multi-span greenhouse; repetitive aisles, sloped ceiling \\
warehouse & racks and aisles; repetitive indoor \\
office & long corridor, small rooms, door transitions, in-place spins \\
stairwell & three stories, U-shaped stairs; identical floors \\
stair\_course & straight and spiral stairs in a courtyard \\
apartment\_building & dog-leg stairs, bare rooms; vertical ambiguity \\
\bottomrule
\end{tabular}
\end{table}

\subsection{Trajectories and carriers}
Each world has one to three trajectories (\cref{tab:sequences}), defined by
waypoints, stops and in-place turns, and densified into 6-DoF trajectories with
the kinematics of one of five carriers: a car (up to \SI{15}{\meter\per\second},
lateral acceleration below \SI{2.5}{\meter\per\second\squared}, body attitude
following the terrain under its wheelbase), a small ground robot (UGV, up to
\SI{3}{\meter\per\second}, in-place spins up to \SI{179}{\degree\per\second}),
a legged robot with a short walking gait and a hand-held rig (both with
gait-induced motion), and a quadruped (\emph{dog} in the sequence
names) whose trot superimposes a \SI{4}{\hertz} body bounce and
roll/pitch/yaw oscillations on top of stair climbing. Every sequence starts
and ends with \SI{3}{\second} at rest for IMU initialization. The
trajectories are stored as keyframes (\SI{20}{\hertz}, \SI{50}{\hertz} for the
quadruped) and interpolated by a $C^2$ spline in position and attitude.

\begin{table*}[t]
\centering
\caption{Sequences: carrier, duration, length, and motion statistics (maximum speed, yaw rate, pitch and roll, and height range).}
\label{tab:sequences}
\footnotesize
\begin{tabular}{@{}llrrrrrr@{}}
\toprule
Sequence & Carrier & Duration & Length & $v_{\max}$ & $\dot\psi_{\max}$ & Pitch/roll & $\Delta z$ \\
 & & [min:s] & [m] & [m/s] & [deg/s] & [deg] & [m] \\
\midrule
apartment\_building/dog\_floors & dog & 7:32 & 273 & 1.4 & 118 & 29/2 & 6.0 \\
apartment\_building/dog\_stairs & dog & 3:20 & 121 & 1.6 & 122 & 29/2 & 6.0 \\
campus/long\_tour & ugv & 15:12 & 1796 & 2.0 & 91 & 2/2 & 5.6 \\
campus/ring\_drive & car & 6:18 & 2792 & 8.0 & 47 & 2/2 & 4.2 \\
forest/trail\_loop & legged & 5:39 & 425 & 1.5 & 22 & 12/11 & 8.0 \\
forest/crossing & ugv & 2:17 & 195 & 1.5 & 5 & 11/10 & 6.4 \\
greenhouse/aisles & ugv & 6:06 & 414 & 1.2 & 82 & 0/0 & 0.0 \\
greenhouse/walk & handheld & 4:13 & 265 & 1.2 & 93 & 1/1 & 0.0 \\
office/walk & handheld & 3:52 & 226 & 1.3 & 96 & 1/2 & 0.1 \\
office/spins & ugv & 2:38 & 120 & 1.0 & 179 & 0/0 & 0.0 \\
open\_field/lap & ugv & 6:32 & 769 & 2.0 & 66 & 5/4 & 7.2 \\
open\_field/zigzag & ugv & 4:42 & 702 & 3.0 & 169 & 5/5 & 7.4 \\
orchard/rows & ugv & 7:52 & 696 & 1.5 & 82 & 3/3 & 4.5 \\
orchard/headland & ugv & 3:41 & 525 & 2.5 & 82 & 2/3 & 5.5 \\
parking\_garage/walk\_down & handheld & 4:01 & 315 & 1.5 & 96 & 10/2 & 6.5 \\
parking\_garage/up\_down & car & 2:51 & 565 & 3.5 & 74 & 9/0 & 6.4 \\
parking\_lot/search & car & 3:42 & 934 & 4.5 & 47 & 0/0 & 0.0 \\
parking\_lot/ugv\_loop & ugv & 5:10 & 569 & 2.0 & 143 & 0/0 & 0.0 \\
rough\_terrain/hill\_loop & ugv & 6:58 & 623 & 1.6 & 72 & 22/19 & 14.1 \\
rough\_terrain/traverse\_fast & ugv & 1:57 & 329 & 3.2 & 35 & 22/13 & 11.4 \\
stair\_course/dog\_course & dog & 2:40 & 110 & 1.5 & 134 & 30/2 & 6.0 \\
stairwell/dog\_floors & dog & 3:48 & 163 & 1.4 & 100 & 30/2 & 7.0 \\
stairwell/legged\_floors & legged & 4:47 & 207 & 1.0 & 87 & 32/1 & 7.0 \\
stairwell/handheld\_stairs & handheld & 2:40 & 118 & 1.2 & 85 & 7/1 & 3.5 \\
tunnel/car\_through & car & 2:28 & 1782 & 15.0 & 32 & 0/0 & 0.0 \\
tunnel/ugv\_inspection & ugv & 5:27 & 798 & 2.5 & 2 & 0/0 & 0.0 \\
urban\_street/car\_loops & car & 5:48 & 2935 & 10.0 & 47 & 0/0 & 0.0 \\
urban\_street/pedestrian & handheld & 8:23 & 671 & 1.5 & 90 & 5/2 & 0.2 \\
warehouse/aisles & ugv & 3:38 & 410 & 2.0 & 82 & 0/0 & 0.0 \\
warehouse/walk & handheld & 3:20 & 260 & 1.4 & 88 & 1/2 & 0.1 \\
\midrule
Total (30 sequences) & & 148 min & 20.1 km & & & & \\
\bottomrule
\end{tabular}
\end{table*}

\subsection{Sensors}
\paragraph{LiDAR}
The main rig is an Ouster OS1-32-like spinning LiDAR (32 uniformly spaced
rings over \SI{45}{\degree}, 1024 columns, \SI{10}{\hertz},
\SIrange{0.5}{60}{\meter}); a Velodyne VLP-16-like rig (16 rings over
\SI{30}{\degree}, 1808 columns, \SIrange{0.4}{100}{\meter}) provides a sparser
second sensor, and an OS1-64-like rig a
denser third one; further sensor models are defined in the repository. Each
point carries its time $t$
relative to the sweep stamp, its ring index, and the semantic class and
object instance of the surface it hit. No intensity is produced since the worlds carry no material model.

\paragraph{Shutter models}
Every sequence is rendered with two shutter models.
In the \emph{global shutter} variant, the whole sweep is cast from the pose at
the sweep start and every point has $t=0$, i.e. there is no motion distortion.
In the \emph{rolling shutter} variant, which reproduces a real rotating
sensor, column $c$ fires at $t_c = c\,T/N_c$ ($T=\SI{0.1}{\second}$,
$N_c$ columns) from the sensor pose at that instant, and its points are
expressed in the sensor frame at their own firing time, as would be measured by a real spinning sensor.
De-skewing is left to the odometry method, with the per-point $t$.

\paragraph{IMU}
The IMU outputs angular velocity and specific force at \SI{200}{\hertz} in its
own frame, mounted with a lever arm of $(0.05, -0.03, 0.45)$\,m from
\texttt{base\_link} (the LiDAR is at $(0, 0, 0.75)$\,m, with identity
rotations). Its readings are computed at the IMU mounting pose by numerical
differentiation of the trajectory spline; \cref{sec:gt} quantifies their
accuracy.

\subsection{Noise levels}
\label{sec:noise}
Three noise levels are rendered. They share the random seed of each sequence,
so that the variants are paired.
\begin{itemize}
\item \emph{clean}: no LiDAR or IMU noise.
\item \emph{noisy}: Gaussian range noise of \SI{2}{\centi\meter} and a MEMS-grade
IMU (\cref{tab:imu}).
\item \emph{real}: the same IMU, and the realistic LiDAR error model below,
with its parameters drawn per sequence from the ranges in
\cref{tab:realnoise}, to vary the error profile across sequences.
The drawn values are stored next to each sequence.
\end{itemize}

The realistic model acts on each ray of each sweep, with $r$ the range and
$\theta$ the incidence angle between the ray and the surface normal.
(i)~The beam has a full divergence $\delta$. Seven sub-rays sample its
footprint: its axis and six rays on the cone of half the divergence, weighted
as a Gaussian beam whose full width at half maximum is $\delta$. Their hits,
in range order, form echoes: a hit joins the previous one if both lie on one
surface, or if they are closer along the beam than the echo separation $L$,
as the pulses they return overlap. An echo's range is the mean of its hits
weighted by their received power, $\cos\theta/r^2$, and the strongest echo
is the return, reported along the axis. On an oblique surface the echo spans
$r\,\delta\tan\theta$ in range, and on an object narrower than the
footprint only the object's depth. At a depth edge, two surfaces closer than
$L$ give a mixed return between both, which stays the same while the
geometry does; farther apart, the return comes from the stronger surface.
(ii)~Gaussian noise with standard deviation
$\sigma(r) = \sqrt{\sigma_0^2 + (k\,r)^2}$ is added, and a uniform term
over the echo's range spread, up to its value at \SI{80}{\degree} of
incidence and at most \SI{0.15}{\meter}: real sensors are not noisier
(\cref{sec:validation}). (iii)~A return needs
$\cos\theta/r^2 \geq 1/R^2$, the received power of a Lambertian surface
seen at normal incidence from the detection range $R$, so that weak returns
at long range and grazing incidence are lost. (iv)~Each ring has fixed
calibration errors drawn once per sequence: elevation and azimuth offsets
(the ray samples the scene along the perturbed direction, but the point is
reported along the nominal one, as with an uncalibrated sensor) and a range
bias. Effect (i) depends on the local geometry and on the scene's depth edges
(vegetation, posts, railings), (iii) on range and incidence, and (iv)
introduces systematic, ring-correlated distortions that do not average out.

\begin{table}[t]
\centering
\caption{Parameter ranges of the realistic LiDAR error model, drawn uniformly per
sequence. The noise, beam and detection ranges bracket two real sensors
(\cref{sec:validation}); the per-ring ranges are assumed.}
\label{tab:realnoise}
\footnotesize
\begin{tabular}{@{}lll@{}}
\toprule
Parameter & Symbol & Range \\
\midrule
Range noise, constant & $\sigma_0$ & \SIrange{0.5}{2}{\centi\meter} \\
Range noise, proportional & $k$ & \numrange{5e-4}{1.5e-3} \\
Beam divergence (FWHM) & $\delta$ & \SIrange{0.1}{0.3}{\degree} \\
Echo separation & $L$ & \SIrange{1.5}{3}{\meter} \\
Detection range (normal incid.) & $R$ & \SIrange{110}{250}{\meter} \\
Per-ring elevation offset (std) & $\sigma_{\mathrm{el}}$ & \SIrange{0}{0.04}{\degree} \\
Per-ring azimuth offset (std) & $\sigma_{\mathrm{az}}$ & \SIrange{0}{0.04}{\degree} \\
Per-ring range bias (std) & $\sigma_{b}$ & \SIrange{0}{1}{\centi\meter} \\
\bottomrule
\end{tabular}
\end{table}

\begin{table}[t]
\centering
\caption{IMU noise in the \emph{noisy} and \emph{real} levels (\SI{200}{\hertz}).
Biases start at zero.}
\label{tab:imu}
\footnotesize
\setlength{\tabcolsep}{2pt}
\begin{tabular}{@{}lll@{}}
\toprule
 & Gyroscope & Accelerometer \\
\midrule
Per-sample noise & \SI{3.7e-3}{\radian\per\second} & \SI{3.2e-2}{\meter\per\second\squared} \\
Noise density & \SI{2.6e-4}{\radian\per\second\per\sqrt\hertz} & \SI{2.3e-3}{\meter\per\second\squared\per\sqrt\hertz} \\
Bias random walk & \SI{2.0e-5}{\radian\per\second\per\sqrt\second} & \SI{1.0e-3}{\meter\per\second\squared\per\sqrt\second} \\
\bottomrule
\end{tabular}
\end{table}

\subsection{The error model against real sensors}
\label{sec:validation}
\begin{figure*}[t]
\centering
\includegraphics[width=\textwidth]{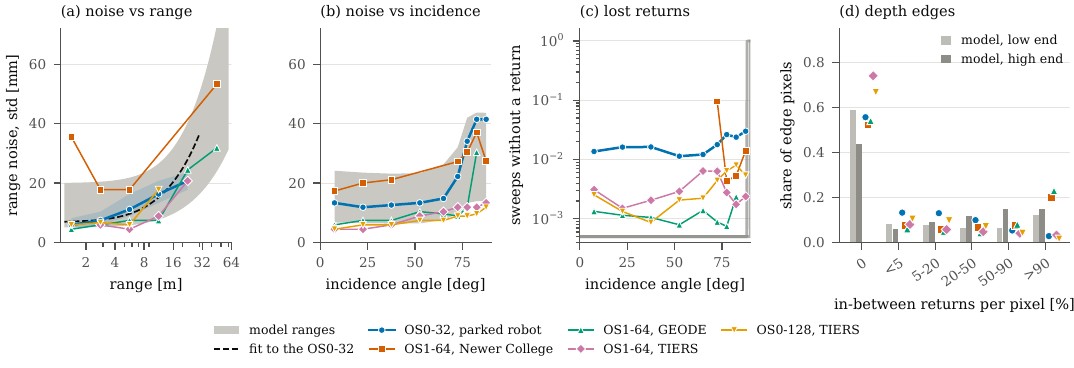}
\caption{The realistic error model against five real Ouster sensors, from
their temporal statistics while still: an OS0-32 on a parked robot, the OS1-64 of Newer College (276 sweeps; in (b) and
(c), incidence from the TLS map), and the OS1-64 of GEODE and the OS1-64 and
OS0-128 of TIERS. Gray: the model over the
parameter ranges of \cref{tab:realnoise}, in (a) at \SI{15}{\degree} of
incidence and in (b) per pixel of the OS0-32. (a) Range noise (median over pixels of the robust temporal standard
deviation, within \SI{30}{\degree} of normal incidence; shaded: interquartile
range of the OS0-32) and the model fitted to the OS0-32 (dashed). (b) Noise
against incidence, 2 to \SI{20}{\meter}. (c) Share of sweeps without a
return, 2 to \SI{20}{\meter}, in log scale: the model loses returns only
beyond its detection range (from \SI{88}{\degree} at \SI{20}{\meter} with
$R=\SI{110}{\meter}$ to \SI{90}{\degree} at \SI{2}{\meter} with
$R=\SI{250}{\meter}$; none below, drawn at the bottom edge); Newer College from
\SI{70}{\degree} only, since below its pixels are on walls with windows, which
drop returns for other reasons. (d) At depth edges, the distribution over
pixels of the share of their returns between the two surfaces: real
(markers), and simulated at the poses of Newer College at the low and high
ends of the ranges.}
\label{fig:validation}
\end{figure*}

We compared the error model with two real Ouster sensors, computing the same
per-pixel statistics on real and simulated sweeps (\cref{fig:validation}).
The first is the OS1-64 (first generation) of the Newer College
dataset \citep{ramezani2020newer}, which stays still for the first
\SI{15.6}{\second} (156 sweeps) of its \texttt{01\_short} sequence, in a site
with a TLS map. The pose of each sweep is the dataset's ground truth refined by
point-to-plane ICP against the TLS cloud (median residual
\SI{10}{\milli\meter}). The reference range of each pixel comes from a ray cast
into the TLS mesh. At the same poses and against the same mesh, the
generator's own LiDAR model renders sweeps with the current error model, at the
ends of the ranges of \cref{tab:realnoise} and for several echo separations
and divergences. Without errors,
it reproduces the reference cast to \SI{0.2}{\milli\meter}.
The second sensor is an OS0-32, with the 32 rings and 1024 columns of the
benchmark's main rig but a \SI{90}{\degree} vertical field of view, on a wheeled
robot that is parked in 30 intervals of \SIrange{10}{52}{\second} over three
recordings in our campus (4937 sweeps). Without a map, its statistics are temporal: the
robust spread over time of each pixel's range (the noise), the incidence angle
from a local plane fit, the share of sweeps with a return and, at depth edges
along a ring, the share of returns that fall between the two surfaces.

To see whether these two units are typical, we computed the same temporal
statistics on three more Ouster units of public datasets, in the intervals in
which their IMU shows them still for at least \SI{10}{\second}: the OS1-64 of
GEODE \citep{chen2025geode} (four intervals in two sequences, 768 sweeps), and
the OS1-64 and OS0-128 of the TIERS dataset \citep{li2022tiers} (two intervals
each, 255 and 256 sweeps); and on 120 more sweeps of the Newer College OS1-64,
still at the start of its \texttt{02\_long} sequence. The three newer units
are the least noisy: within \SI{30}{\degree} of normal incidence, 4 to
\SI{9}{\milli\meter} up to \SI{8}{\meter} and 7 to \SI{18}{\milli\meter} at
8--\SI{16}{\meter}, against 6 to \SI{16}{\milli\meter} up to \SI{16}{\meter}
for the OS0-32, and \SI{18}{\milli\meter} at 2--\SI{8}{\meter} and
\SI{36}{\milli\meter} at 1--\SI{2}{\meter} for the first-generation OS1-64.

Four properties of the two sensors shaped the model.
First, real sensors keep returning at grazing incidence. Within
\SI{20}{\meter}, both return in 97 to \SI{100}{\percent} of the sweeps up
to 85--\SI{90}{\degree} (on Newer College, with the incidence from the TLS
normals, \SI{99}{\percent} at 80--\SI{85}{\degree} and
\SI{98}{\percent} at 85--\SI{88}{\degree}), so the model has no incidence
cutoff, only the detection range.
Second, the noise grows with range: on the OS0-32, from
\SI{6}{\milli\meter} at 1--\SI{2}{\meter} to \SI{21}{\milli\meter} at
16--\SI{32}{\meter}. A least-squares fit of the noise terms over 53 bins of
range and incidence gives $\sigma_0=\SI{6.8}{\milli\meter}$,
$k=\num{1.14e-3}$ and $\delta=\SI{0.28}{\degree}$ (relative RMS residual
0.21). The OS1-64 is noisier, with \SI{55}{\milli\meter} at
32--\SI{64}{\meter}.
Third, real mixed returns persist on the same pixels while the sensor is
still: \SI{52}{\percent} of the edge pixels of the OS1-64 and
\SI{56}{\percent} of those of the OS0-32 never return between the two
surfaces, against 2 to \SI{4}{\percent} if mixed returns are drawn at random
between neighbor rays.
Fourth, real noise stops growing with the footprint: beyond about
\SI{80}{\degree} of incidence, neither sensor is noisier than about
\SI{40}{\milli\meter}, so the footprint's range spread is capped.

The model (\cref{sec:noise}) samples the footprint, so that mixed
returns follow the geometry, and its ranges (\cref{tab:realnoise}) bracket
the two sensors they were set from (\cref{fig:validation}). They cover the
noise against range and incidence, and returns up to \SI{88}{\degree} of
incidence within \SI{20}{\meter}. The three newer units sit at the low end of
the noise ranges (at most \SI{1.4}{\milli\meter} below it), and the
first-generation OS1-64 above the high end at 1--\SI{2}{\meter}. Below
\SI{88}{\degree}, the model loses no returns within \SI{20}{\meter}, where the
real units lose 0.1 to \SI{3}{\percent} at every incidence (the OS0-32 1 to
\SI{3}{\percent}, the others 0.1 to \SI{1.5}{\percent}, apart from the walls
with windows of Newer College). At the depth edges of Newer College, the
model's ranges give a share of returns between the two surfaces of 5.4 to
\SI{18.8}{\percent}, against \SI{10.7}{\percent} for the OS1-64, with the
edges taken from the mesh. Real mixing reaches depth gaps of 1 to
\SI{4}{\meter}; the echo separation $L$ is drawn inside this span
(\SIrange{1.5}{3}{\meter}). The share of edge pixels that never mix is 44 to
\SI{59}{\percent}, against 52 to \SI{74}{\percent} for the five units. At
extreme grazing (85--\SI{90}{\degree}) the model is noisier than most units:
at the poses of Newer College it gives \SIrange{48}{61}{\milli\meter}
against \SI{28}{\milli\meter} measured there, and the two TIERS units (12 to
\SI{13}{\milli\meter}) lie below its range at their pixels
(\SIrange{17}{54}{\milli\meter}); only the OS0-32 (\SI{42}{\milli\meter})
lies within it (\cref{fig:validation}b). The per-ring biases could not
be separated from the errors of the mesh.

\subsection{Variants}
Each trajectory is released in nine variants named
\texttt{<rig>-<noise>-<shutter>}: the OS1-32 rig at the three noise levels
with both shutter models, the VLP-16 rig with the \emph{noisy} level and
rolling shutter, and the OS1-64 rig with the \emph{noisy} and \emph{real}
levels and rolling shutter. The \emph{real} variants are the primary
benchmark; the others serve as controlled ablations. In total, the release
holds 270 rendered sequences, nine renders of the \SI{20.1}{\kilo\meter} and
\SI{148}{\minute} of trajectories (\SI{181}{\kilo\meter} and \SI{22.2}{\hour} of
sensor data), 798~thousand LiDAR sweeps with 25.2~billion points,
16.0~million IMU samples, and \SI{272}{\giga\byte} of compressed data. The realistic error model
removes \SI{0.7}{\percent} of the returns of the noise-free sweeps on average: the weak returns at long
range and grazing incidence that fall below its detection range. It adds no
returns, as the beam's footprint is sampled only around rays whose axis hits.

\subsection{Data format and tools}
Each sequence variant is an MRPT rawlog (a compressed, time-ordered stream of
point-cloud, IMU and odometry observations, read with the MRPT
libraries), plus TUM-format ground truth for \texttt{base\_link}
and the LiDAR at every sweep, and the \SI{200}{\hertz} ground truth. The
repository provides the world and trajectory generators, the rendering
scripts, an IMU consistency checker, a tool that builds the exact reference
map of a sequence (or the map implied by an estimated trajectory, for map
metrics), and the evaluation scripts used in this paper. All the LiDAR
odometry methods evaluated here read the rawlogs directly through MOLA.
The world frame has $z$ up, with gravity of \SI{9.81}{\meter\per\second\squared}
along $-z$; every sequence starts at $t=\SI{100}{\second}$, and each sweep is
stamped at its start. The files also carry a planar wheel odometry at
\SI{10}{\hertz}, derived from the ground truth (with \SI{1}{\percent}
relative noise in the \emph{noisy} variants) for every carrier, including
those where it has no physical meaning (hand-held, legged and quadruped); it
is not used in this paper.

So that the data does not depend on a binary format or a particular library,
every variant is also released as a plain-text copy (\texttt{text/} in the
data repository): one LiDAR point per line
(\texttt{t\_ns x y z ring label instance}, the sweeps one after another), a
table of sweeps with their stamps and point counts, the IMU and odometry
readings, the extrinsics as YAML, and the same TUM ground truth. The tables
are compressed with xz, which the standard library of Python and common
archivers read, and a short reader that needs only Python and NumPy is
included. The copy holds the same samples as the rawlogs; only the point
coordinates are rounded to \SI{1}{\milli\meter} (\SI{0.3}{\milli\meter} RMS).
It takes \SI{101}{\giga\byte}, and its sweep and point counts were checked
against those of every rawlog.

Every point is labeled with the semantic class of the surface it hit (outdoor
classes reuse the SemanticKITTI label ids \citep{behley2019semantickitti};
indoor ones are added) and with an object instance id, for use by learned
perception methods. The labels come from the same ray casts as the points, so
they are exact, and adding them does not change any point. Returns that mix
two surfaces at a depth edge are labeled as outliers, so the labels reveal the
injected mixed pixels; they are not given to the odometry methods.

\section{Ground Truth}
\label{sec:gt}

\paragraph{Trajectories}
The ground truth is the trajectory used to drive the simulator. We release
it as the spline keyframes (TUM text files in the repository), as the pose of
\texttt{base\_link} and of the LiDAR at every sweep stamp, and densely at
\SI{200}{\hertz}. Since methods stamp their output at the sweep start, middle
or end, all our metrics evaluate the spline at each estimate's own timestamp,
so that no association or interpolation error enters the score. The
extrinsics are exact and there is a single clock, so there is no calibration
or synchronization error either.

\paragraph{Maps}
The world geometry is known analytically. For evaluation, we render a
reference point map per sequence: the noise-free, global-shutter sweeps placed
at the true poses, accumulated into a voxel grid (\SI{10}{\centi\meter}
centroids by default). This map contains the surfaces that the sensor
observed along the trajectory, up to its voxelization, so completeness is
meaningful. A map built from
an estimated trajectory is computed the same way (every sweep, the same
voxels) from the variant's own sweeps, each point placed with the estimated
trajectory, after the $SE(3)$ alignment of the APE, interpolated at its
firing time, i.e., de-skewed by the estimate itself.

\paragraph{Verification}
We check that the rendered data agrees with the ground truth with three
independent tests. (1)~\emph{IMU}: the noise-free angular velocity and
specific force are recomputed at the IMU mounting pose from the spline, with
its analytic derivatives and lever arm, and compared sample by sample. On
wheeled sequences (\texttt{warehouse/aisles}, \texttt{urban\_street/car\_loops})
the RMS error is below \SI{2e-7}{\radian\per\second} for the gyroscope and
below \SI{5e-6}{\meter\per\second\squared} for the accelerometer; with the
quadruped trot (\texttt{stairwell/dog\_floors}), whose \SI{4}{\hertz} bounce
stresses the finite differences, it is \SI{3.4e-5}{\radian\per\second} and
\SI{1.5e-3}{\meter\per\second\squared}, against a signal of
\SI{2}{\meter\per\second\squared} RMS. (2)~\emph{Shutter and per-point time}:
the noise-free rolling-shutter sweeps, de-skewed with the true trajectory at
each point's time, reproduce the global-shutter map to the floor set by the
voxelization (\texttt{forest/crossing}: median distance between voxel
centroids \SI{1.9}{\centi\meter}, \SI{97.6}{\percent} of them within
\SI{10}{\centi\meter}), while the same sweeps without de-skew do not
(\SI{4.1}{\centi\meter}, \SI{86.4}{\percent}). (3)~\emph{End to end}: on
noise-free, global-shutter data, LiDAR odometry reaches millimeter-level APE
on the structured worlds (MOLA-LIO: 1.3 and \SI{1.7}{\milli\meter} on
\texttt{greenhouse/aisles} and \texttt{warehouse/aisles}; GLIM and DLIO: 6 to
\SI{7}{\milli\meter} on their best sequences), which bounds any residual
inconsistency between sensor data, extrinsics and ground truth.

\paragraph{Observability}
\label{sec:observability}
Exact ground truth also tells whether a scene constrains the motion at all,
independently of any method. Each clean global-shutter OS1-32 sweep is placed
at its true pose and matched to the reference map, whose local plane fits give
each point a surface normal. The Fisher information of point-to-plane
registration against that map, with the \SI{2}{\centi\meter} range noise of
the \emph{noisy} level, gives the Cram\'er-Rao bound of the sweep's
translation, with its rotation also unknown
\citep{censi2007achievable,zhang2016degeneracy}: a lower bound for any
estimator that registers the sweep, since it assumes known correspondences
and independent noise. Each fifth sweep is also shifted along the direction
of travel, counting the points that end farther than \SI{10}{\centi\meter}
from the map, which shows ambiguities that the local bound cannot. A
one-dimensional Kalman filter along the direction of travel, with the
accelerometer of \cref{tab:imu} and at each sweep a displacement measured
with that bound, gives the drift of an ideal LiDAR-inertial odometry.
\Cref{tab:observability} gives the results for the two degenerate scenes
(open field and tunnel, four sequences), the campus and three others.
No sequence is unobservable: in every sweep the information has full rank,
and the bound along the direction of travel stays below \SI{11}{\milli\meter}
in the tunnel and \SI{25}{\milli\meter} in the open field. The ideal
odometry drifts by 7 and \SI{23}{\centi\meter} over the tunnel car and UGV
runs, and by 0.59 and \SI{0.33}{\meter} over the open-field lap and zigzag (one
standard deviation). What makes these scenes hard is how few points carry the
information. Inside the straight sections of the tunnel, on the UGV run,
\SI{95}{\percent} of the information along the axis comes from
\SI{0.6}{\percent} of the points, those on surfaces that face the axis (the
sides of the emergency niches and the ends of lamps, signs and fans): 23 per
sweep in the median and, in the worst sweep, one. The bound along the axis is
about 23 times that across it, and shifting a sweep of the UGV run by
\SI{10}{\meter}, the spacing of the lamps, moves as few as 5 of its
\num{32500} points off the map, against at least \num{4000} on the urban
street. In the open field both horizontal directions are weak, and a shift of
\SI{0.5}{\meter} moves about \SI{1}{\percent} of the points; but the relief of
the ground moves most of them at \SI{5}{\meter} and more, so the difficulty
there is a weak local constraint rather than an ambiguity. The campus is well constrained (a bound of at most
\SI{2}{\milli\meter}, at least 250 points moved off the map by a
\SI{0.5}{\meter} shift, and an ideal drift of 6 to \SI{8}{\centi\meter}), so
the 2 to \SI{69}{\meter} of APE that the methods reach on its ring road
(\cref{sec:difficulty}) come from the methods, not from the data. The
failures in these scenes (\cref{sec:difficulty}) are therefore not imposed by
a lack of information in the data: an estimator has to keep, and trust, a few
dozen points among tens of thousands. The bound does not include the per-ring
biases and the range-dependent noise of the \emph{real} level.

\begin{table}[t]
\centering
\caption{Observability of the clean OS1-32 sweeps at their true poses, independent of any method: Cram\'er-Rao bound of the translation along and across the direction of travel (range noise \SI{2}{\centi\meter}; along: median and maximum over the sweeps, across: median), and the points that end farther than \SI{10}{\centi\meter} from the reference map when a sweep is shifted along the direction of travel (median and minimum over every fifth sweep; \num{16000} to \num{33000} points per sweep). $^\ast$Inside the tunnel, farther than \SI{70}{\meter} from the portals.}
\label{tab:observability}
\footnotesize
\setlength{\tabcolsep}{3pt}
\begin{tabular}{@{}lrrrr@{}}
\toprule
 & \multicolumn{2}{c}{Bound [mm]} & \multicolumn{2}{c}{Points off the map} \\
\cmidrule(lr){2-3}\cmidrule(l){4-5}
Sequence & along & across & \SI{0.5}{\meter} & \SI{10}{\meter} \\
\midrule
urban street, car loops & 0.3 / 0.5 & 0.21 & 5324 / 2392 & 7861 / 4018 \\
greenhouse, aisles & 0.3 / 0.4 & 0.16 & 8104 / 3069 & 8080 / 4190 \\
rough terrain, hill loop & 1.3 / 2.3 & 1.24 & 2677 / 339 & 17392 / 6903 \\
campus, ring drive & 0.9 / 1.9 & 0.60 & 1904 / 514 & 15529 / 2744 \\
campus, long tour & 0.7 / 2.0 & 0.53 & 2672 / 258 & 17380 / 665 \\
open field, lap & 7.8 / 24.2 & 4.88 & 137 / 14 & 14426 / 1788 \\
open field, zigzag & 5.1 / 16.7 & 3.09 & 168 / 25 & 14802 / 1995 \\
tunnel, car$^\ast$ & 3.2 / 6.9 & 0.14 & 85 / 22 & 599 / 15 \\
tunnel, UGV$^\ast$ & 3.7 / 10.5 & 0.17 & 52 / 5 & 438 / 5 \\
\bottomrule
\end{tabular}
\end{table}

\section{Evaluation Protocol}
\label{sec:protocol}

\paragraph{Methods}
We evaluate eight configurations of seven open-source methods, all run through
their MOLA wrappers \citep{blanco2025mola} from the same rawlog files:
MOLA-LO (GICP pipeline \citep{segal2009gicp}, default parameters, with the IMU
used for de-skew, gravity alignment and a gyroscope prior on the rotation of
each registration) with its simple constant-velocity state estimator, and
MOLA-LIO, the same front end with the factor-graph smoother and IMU
preintegration \citep{forster2017preintegration}; the
LiDAR-only KISS-ICP \citep{vizzo2023kissicp} and the continuous-time
Traj-LO \citep{zheng2024trajlo}; and the LiDAR-inertial
DLIO \citep{chen2023dlio}, FAST-LIO2 \citep{xu2022fastlio2}, the odometry
stage of GLIM \citep{koide2024glim} (without its mapping back end) and
EllipseLIO \citep{ellipselio}, each built from the latest upstream code of
the method (November 2026; KISS-ICP v1.3.0; the commits are listed in the
repository). Only KISS-ICP and Traj-LO are LiDAR-only. No loop closure is
used, and no method uses the wheel odometry stream present in the files.

\begin{table*}[t]
\centering
\caption{Evaluated configurations. De-skew is enabled on rolling-shutter data
only. The last column is the part of the map each method registers against
with its default parameters.}
\label{tab:methods}
\footnotesize
\setlength{\tabcolsep}{2pt}
\begin{tabular}{@{}lllll@{}}
\toprule
Method & Registration & IMU use & De-skew & Map kept \\
\midrule
MOLA-LO & GICP, local map & de-skew, gravity, gyro & IMU & \SI{100}{\meter} radius \\
MOLA-LIO & same & + preintegration & IMU & same \\
KISS-ICP & pt.-to-pt.\ ICP & none & const.\ vel. & max.\ range \\
Traj-LO & continuous-time & none & trajectory & all \\
DLIO & GICP, keyframes & observer & IMU & kNN keyframes \\
FAST-LIO2 & pt.-to-plane, iEKF & tight & IMU & \SI{1}{\kilo\meter} box \\
GLIM & GICP, smoother & tight & IMU & last 100 scans \\
EllipseLIO & ellipsoids, iESKF & tight & IMU & all \\
\bottomrule
\end{tabular}
\end{table*}

\paragraph{Configuration}
Each method is first run with its published default parameters, one parameter
set for all 30 sequences, changing only what describes the sensor: the
extrinsics; the minimum and maximum ranges (and KISS-ICP's map voxel size,
which its pipeline derives from the maximum range); the rings, vertical field
of view and rates where a method asks for them; and the shutter-dependent
switch:
\begin{itemize}
\item global shutter: de-skew disabled and per-point times ignored;
\item rolling shutter: de-skew enabled with the per-point times (IMU-based for
the LIO methods and MOLA, constant velocity for KISS-ICP, the continuous-time
trajectory for Traj-LO).
\end{itemize}
The defaults are those of each method's repository (FAST-LIO2: its
spinning-LiDAR launch parameters; GLIM: its odometry configuration).
EllipseLIO ships IMU noise values per sensor only, so it uses its
configuration for a comparable hand-held spinning LiDAR (Oxford Spires). Two
departures remain: FAST-LIO2 receives every point, since its default
decimation assumes a different point order, and MOLA's default fusion of
the wheel odometry is turned off. No method is given the simulated IMU's noise
parameters in this configuration.
Traj-LO requires per-point times and is evaluated on rolling shutter only.
All runs are single-threaded, each pinned to one core, with the options of
the MOLA wrappers that make the results repeatable: DLIO and FAST-LIO2 rebuild their map structure (submap, ikd-Tree)
synchronously instead of in the background, and EllipseLIO's iterated update
has no wall-clock budget. 
Runs that need more than
\SI{16}{\giga\byte} of memory are stopped, marked in \cref{fig:heatmap} and
left out of the statistics; only Traj-LO, which keeps its whole map, did hit this limit.

\paragraph{Tuned configurations}
Defaults favor the data their authors tuned them on. To separate the method
from the tuning effort, every method is also tuned with the same protocol and
budget: up to four of its key parameters, each either at its default or at
one alternative (listed with the evaluation code), that is, at most 16 configurations per
method. All are run on a training split of five worlds chosen before any
tuning result (office, apartment building, parking lot, orchard and tunnel:
10 sequences, indoor, outdoor, natural and degenerate), at the \emph{noisy}
level with both shutters. The configuration with the lowest mean APE,
expressed as a percentage of the path length and capped at the
\SI{10}{\percent} failure threshold (a failed run counts as the cap), is
kept; ties go to the one closer to the defaults. It is one configuration per
method for every test sequence and variant, reported on the 20 sequences of
the other ten worlds; no parameter is tuned per sequence.

\paragraph{Trajectory metrics}
For each run we report the absolute pose error (APE), the RMSE of the
position error after an $SE(3)$ Umeyama alignment \citep{umeyama1991}, and
the relative translation error over \SI{10}{\meter} segments (RTE,
\si{\percent} of the segment length), as in the KITTI odometry
benchmark \citep{geiger2012kitti} (see
also \citep{sturm2012tum,zhang2018tutorial}). Since an $SE(3)$ fit can hide a
tilted estimate, we also give the APE after a fit in yaw and translation only
(APE$_4$), with each estimate first rotated so that its first attitude is the
true one (MOLA, FAST-LIO2 and the LiDAR-only methods write their trajectory in
the frame of the first pose), and the RMS tilt error, the angle between the
true and estimated gravity directions in the body frame. The released result
files also give the rotation APE, the RTE over \SI{50}{\meter}, the relative
rotation error and the end-point drift. A run
is a \emph{failure} if its output covers less than \SI{90}{\percent} of the
sequence, if the ratio of its path length to the true one is outside 0.8 to
1.25, or if its APE exceeds \SI{10}{\percent} of the distance traveled. Path lengths are measured between the poses at every
\SI{1}{\meter} of the true path, so that pose jitter of an otherwise accurate
estimate does not count as distance. Stamp conventions were checked by
fitting the APE-optimal constant time offset to every noise-free run with an
APE under \SI{10}{\centi\meter}: applying the fitted offsets would lower the
APE by less than \SI{5}{\milli\meter}, so no offset is applied.

\paragraph{Map extent}
The methods keep maps of very different extents (\cref{tab:methods}). One
that keeps the whole map re-registers against it when it revisits a place,
an implicit loop closure that lowers its APE but not its local drift. We
therefore report RTE next to APE, and quantify the effect by changing the
map extent of MOLA (\SI{100}{\meter} to \SI{1}{\kilo\meter}) and of FAST-LIO2
(\SI{1}{\kilo\meter} to \SI{200}{\meter}) (\cref{sec:map-extent}).

\paragraph{Statistics}
Effects are measured as paired ratios: the median, over the sequences where
both runs succeed, of the APE of a run over that of its pair (another
variant, configuration or method on the same sequence). Their 95\,\%
intervals are percentile bootstrap intervals over the 15 worlds
(\num{10000} resamples, each world with all the runs of its sequences, since
the sequences of a world share its geometry); they express how much a result
depends on the choice of sequences, not on the variability of real
environments. Success curves give, for every threshold, the share of runs
whose APE is below that fraction of the path length; failed runs never count
as successful.

\paragraph{Map metrics}
Between the estimated map and the reference map we compute the accuracy
(mean, median and 95th percentile of the distance from each estimated voxel to
the reference), the completeness (mean distance from the reference to the
estimate), their average (Chamfer distance), and the precision, recall and
F-score at \SI{10}{\centi\meter} and \SI{20}{\centi\meter}. Distances are
capped at \SI{2}{\meter}. The \SI{10}{\centi\meter} threshold equals the voxel
size of the maps, so on the \emph{real} variants even the true poses do not
reach an F-score of one there (\cref{tab:maps}).

\section{Results}
\label{sec:results}

\subsection{Main benchmark}
\begin{figure}[t]
\centering
\includegraphics[width=\columnwidth]{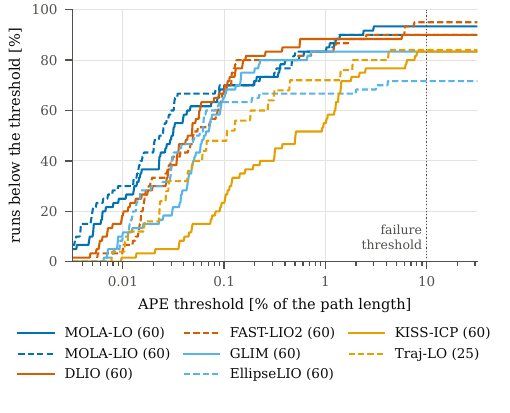}
\caption{Success curves on the two OS1-32 \emph{real} variants: share of the
runs whose APE is below a fraction of the path length (number of runs in
parentheses; Traj-LO: its rolling-shutter runs that fit in memory). Failed
runs never count; the dotted line is the threshold of the failure rule.}
\label{fig:success}
\end{figure}

\begin{figure*}[t]
\centering
\includegraphics[width=\textwidth]{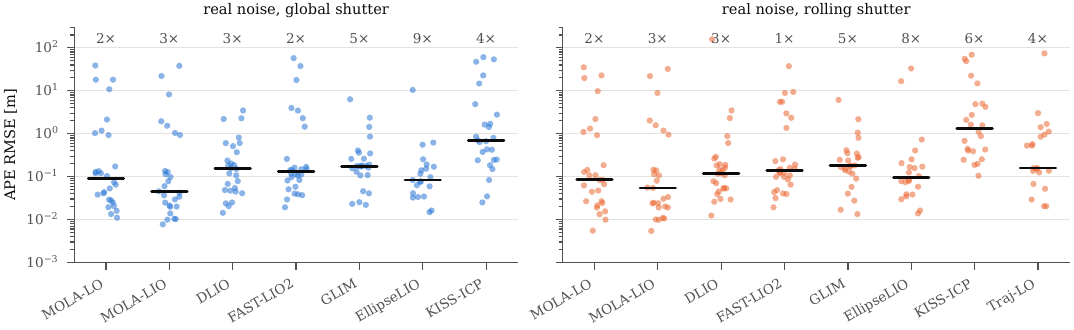}
\caption{APE per sequence (dots) and median (bar) of the successful runs on
the \emph{real} noise level; the number of failures ($N\times$) is printed at the top.}
\label{fig:ape_strip}
\end{figure*}

\begin{figure*}[t]
\centering
\includegraphics[width=\textwidth]{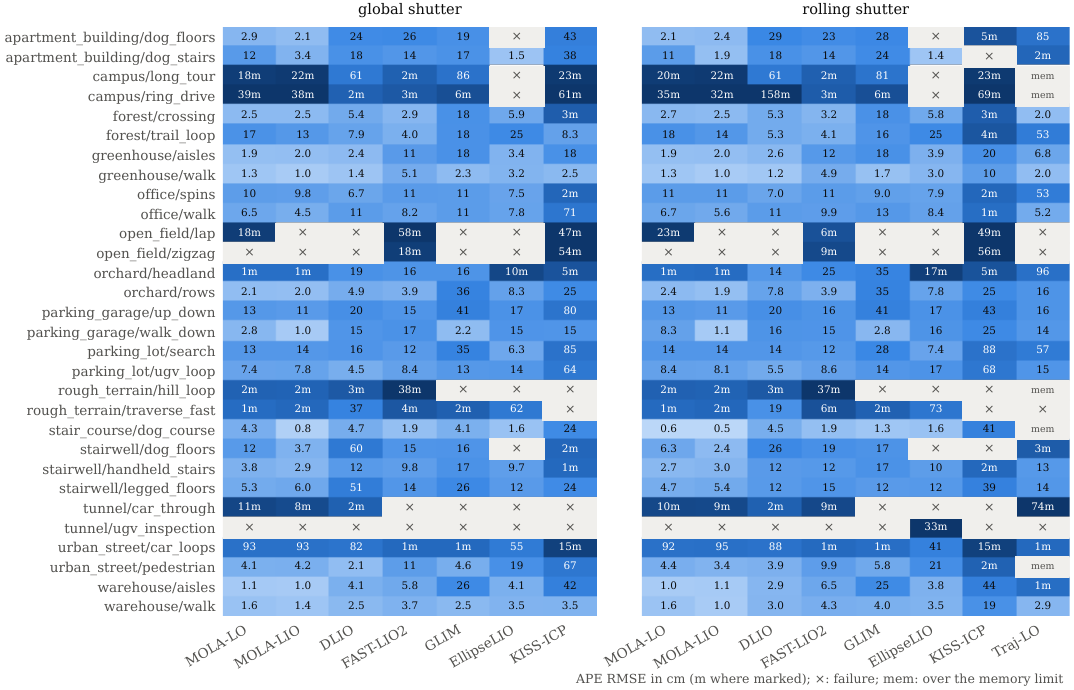}
\caption{APE (RMSE) per sequence and method on the \emph{real} noise level,
both shutter models. Each cell gives the APE in centimeters (or in meters where denoted with ``m''); $\times$: failure; mem: run stopped at the memory
limit.}
\label{fig:heatmap}
\end{figure*}

\begin{figure*}[t]
\centering
\includegraphics[width=\textwidth]{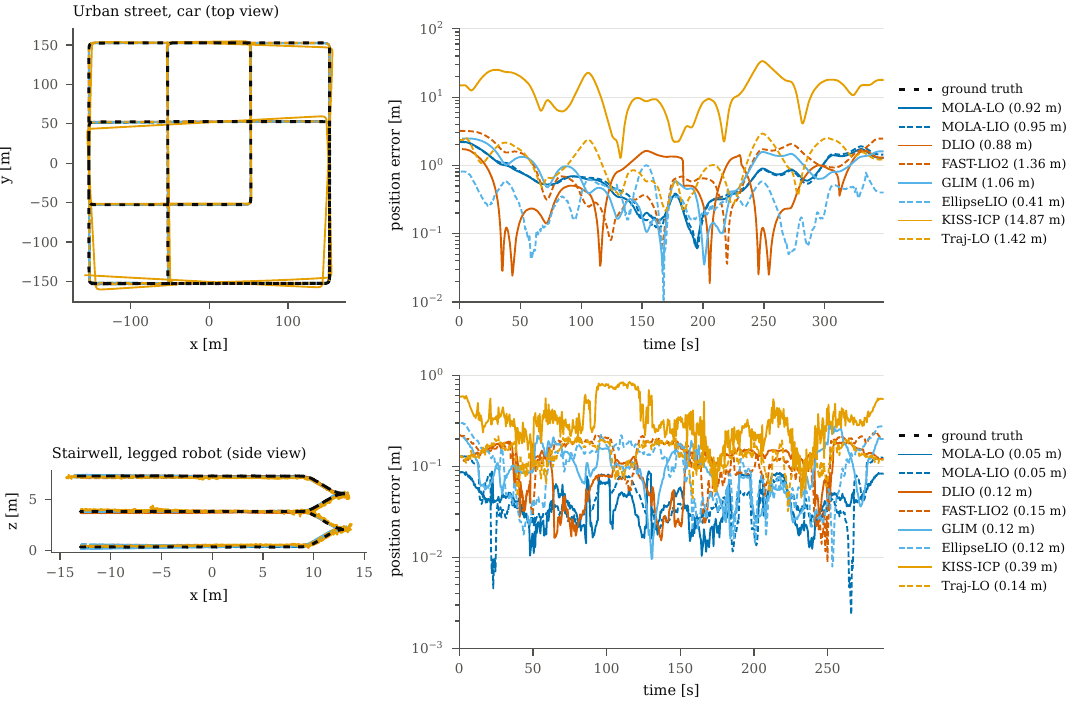}
\caption{Two runs on the \emph{real} noise level with rolling shutter. Left:
ground truth (black, dashed) and the estimates after SE(3) alignment. Right: norm of the
position error after that alignment, whose RMSE is the APE given in the
legend; since the alignment is fitted to the whole trajectory, the error is
not zero at the start. Top: urban car loop (\SI{2.9}{\kilo\meter}); bottom: legged robot over
three floors of the stairwell.}
\label{fig:traj_examples}
\end{figure*}

\Cref{tab:main_summary} summarizes the \emph{real} noise level, the primary
benchmark; \cref{fig:ape_strip} shows the spread over the sequences, and
\cref{fig:heatmap} gives every run (the values are released with the
evaluation code). With their default
parameters, MOLA-LIO has the lowest median APE with both shutter models
(\SI{0.045}{\meter} and \SI{0.054}{\meter}), followed by EllipseLIO and
MOLA-LO (0.083 to \SI{0.094}{\meter}), and FAST-LIO2 fails on the fewest
sequences (2 and 1 of 30), followed by MOLA-LO (2) and by MOLA-LIO and DLIO
(3). EllipseLIO fails on 9 and 8 sequences; 53 of its 71 failures over the
nine variants are in the open field, the tunnel, the campus and the
rough-terrain loop, and 16 of the other 18 on stairs and ramps. FAST-LIO2,
DLIO and GLIM follow (0.12 to \SI{0.18}{\meter}), as does Traj-LO on rolling
shutter, and KISS-ICP is an order of magnitude less accurate. Medians over
different sets of sequences compare methods only loosely: on the sequences
that both complete, MOLA-LIO's APE is lower than FAST-LIO2's, GLIM's and
EllipseLIO's by a median factor of 2.1 to 2.5 (95\,\% intervals from 1.2 to
4.7) and than MOLA-LO's by \SI{7}{\percent} (interval 1.01 to 1.14), but it
cannot be told apart from DLIO's (1.74, interval 0.96 to 3.45). The failure counts depend on
the \SI{10}{\percent} threshold, and \cref{fig:success} gives the share of
runs below every threshold: MOLA-LIO and MOLA-LO have the most runs below
\SI{0.05}{\percent} of the path (67 and \SI{62}{\percent}), DLIO between 0.2
and \SI{0.5}{\percent} (82 and \SI{85}{\percent}), and FAST-LIO2 at the
failure threshold (\SI{95}{\percent}); KISS-ICP stays below every other
method up to \SI{1}{\percent}, and EllipseLIO levels off at
\SI{72}{\percent}. Per sequence, MOLA-LIO has the lowest APE on 13 of the
29 (global) and 9 of the 30 (rolling) sequences that at least one method
completes, and is within 1.5 times the best on 16 and 18; DLIO is the best on
7 and 6, and MOLA-LO on 3 and 5. The APE spans four orders of magnitude
across sequences, from under a centimeter on the stair course to tens of
meters on the long and degenerate ones (\cref{sec:difficulty}).
The fit in yaw and translation, which cannot hide a tilt, raises the APE of
each run, by a median of 6 to \SI{24}{\percent} for the LiDAR-inertial methods
but GLIM (38 to \SI{49}{\percent}), and of 17 to \SI{64}{\percent} for the
LiDAR-only ones (the medians in \cref{tab:main_summary} grow more, as the fit
raises the smallest errors the most in relative terms);
MOLA-LIO and MOLA-LO stay first and second, and EllipseLIO falls back to the
level of DLIO and FAST-LIO2 (\SIrange{0.14}{0.18}{\meter}). The median tilt
error is 0.14 to \SI{0.20}{\degree} for MOLA, 0.26 to \SI{0.48}{\degree} for
the other LiDAR-inertial methods, \SI{0.9}{\degree} for Traj-LO and 3.6 to
\SI{3.9}{\degree} for KISS-ICP.
\Cref{fig:traj_examples} shows two runs. On the \SI{2.9}{\kilo\meter} urban
car loop every method completes, but the drift of KISS-ICP grows to an APE
of \SI{15}{\meter}, against 0.4 to \SI{1.4}{\meter} for the others. In the
stairwell, MOLA-LO and MOLA-LIO keep an APE of \SI{5}{\centi\meter} over the
three floors, the other LiDAR-inertial methods and Traj-LO 12 to
\SI{15}{\centi\meter}, and KISS-ICP \SI{39}{\centi\meter}.
\begin{table}[t]
\centering
\caption{Summary over the 30 sequences: failures out of the runs, and median over the successful runs of the APE [m], the APE after a yaw and translation fit with the first attitude given (APE$_4$ [m]), the RMS tilt error [\si{\degree}] and the RTE over 10~m [\%]. Runs stopped at the memory limit (Traj-LO only) are left out.}
\label{tab:main_summary}
\footnotesize
\setlength{\tabcolsep}{3pt}
\begin{tabular}{@{}llrrrrr@{}}
\toprule
Variant & Method & Fail & APE & APE$_4$ & Tilt & RTE$_{10}$ \\
\midrule
os1\_32-real-global & MOLA-LO & 2/30 & 0.089 & 0.112 & 0.16 & 0.21 \\
 & MOLA-LIO & 3/30 & 0.045 & 0.080 & 0.14 & 0.12 \\
 & DLIO & 3/30 & 0.151 & 0.168 & 0.33 & 0.26 \\
 & FAST-LIO2 & 2/30 & 0.129 & 0.162 & 0.42 & 0.25 \\
 & GLIM & 5/30 & 0.176 & 0.276 & 0.48 & 0.46 \\
 & EllipseLIO & 9/30 & 0.083 & 0.164 & 0.29 & 0.20 \\
 & KISS-ICP$^\dagger$ & 4/30 & 0.688 & 1.310 & 3.64 & 1.94 \\
\midrule
os1\_32-real-rolling & MOLA-LO & 2/30 & 0.084 & 0.105 & 0.20 & 0.19 \\
 & MOLA-LIO & 3/30 & 0.054 & 0.070 & 0.16 & 0.13 \\
 & DLIO & 3/30 & 0.119 & 0.142 & 0.38 & 0.26 \\
 & FAST-LIO2 & 1/30 & 0.142 & 0.182 & 0.44 & 0.24 \\
 & GLIM & 5/30 & 0.180 & 0.331 & 0.44 & 0.44 \\
 & EllipseLIO & 8/30 & 0.094 & 0.173 & 0.26 & 0.21 \\
 & KISS-ICP$^\dagger$ & 6/30 & 1.340 & 1.705 & 3.85 & 3.03 \\
 & Traj-LO$^\dagger$ & 4/25 & 0.159 & 0.587 & 0.87 & 0.53 \\
\bottomrule
\multicolumn{7}{@{}l}{\scriptsize $^\dagger$LiDAR only.}\\
\end{tabular}
\end{table}

\subsection{Tuned configurations}
\label{sec:tuned}
On the training split, tuning pays off for the methods whose defaults are
coarse for this sensor: finer scan and map voxels lower
FAST-LIO2's score from \SI{1.57}{\percent} to \SI{0.45}{\percent} of the path
length and remove its three failures, and the finer iVox halves GLIM's;
KISS-ICP, EllipseLIO (with the true IMU noise) and Traj-LO gain less. MOLA-LO,
MOLA-LIO and DLIO gain nothing: all configurations of the two MOLA methods,
and DLIO's best ones, score within \SI{0.03}{\percent} of the defaults, a
score dominated by the same two failures in the tunnel. The chosen
configurations are MOLA-LO with denser ICP input, MOLA-LIO with denser
ICP input and the ICP covariance as the LiDAR increment noise, KISS-ICP with
the finer voxel, Traj-LO with the denser input, DLIO with the larger submap,
FAST-LIO2 with both finer voxels, GLIM with the finer iVox and the true IMU
noise, and EllipseLIO with the true IMU noise and bias random walk.

On the 20 test sequences (\cref{tab:tuned}, and over all nine variants in the
last column of \cref{tab:effects}), the tuned FAST-LIO2 and GLIM lower their
APE by a median \SI{44}{\percent} and \SI{28}{\percent} of the paired runs
(95\,\% intervals 9 to 53 and 13 to \SI{54}{\percent}; 127 of 165 and 125 of
155 runs improve), which brings GLIM (\SI{0.061}{\meter}
and \SI{0.076}{\meter} on the \emph{real} variants) to the level of MOLA-LO
(\SI{0.058}{\meter} and \SI{0.053}{\meter}) and FAST-LIO2
(\SI{0.076}{\meter} and \SI{0.089}{\meter}) ahead of DLIO; the tuned
MOLA-LIO has \SI{0.038}{\meter} and \SI{0.036}{\meter}. The paired APE of the
other methods changes by \SI{8}{\percent} or less, with intervals that
include no change (the medians of
\cref{tab:tuned} are over different sets of successful runs). Tuning does not
make the methods more robust: over 180 test runs it adds failures to MOLA-LO
(4 to 11), MOLA-LIO (7 to 11) and FAST-LIO2 (3 to 14), nearly all in the open
field, one or two to DLIO and GLIM, none to Traj-LO, and removes them only for
EllipseLIO (51 to 36) and KISS-ICP (41 to 29). FAST-LIO2, which no longer
fails on the training split, fails with its finer voxels on both open-field
runs of every OS1-32 variant.

\begin{table}[t]
\centering
\caption{Defaults (D) and the tuned configuration (T) on the 20 test sequences of the \emph{real} noise level: failures, and median APE [m] of the successful runs (over different sets of sequences for D and T; \cref{tab:effects} gives the paired ratios).}
\label{tab:tuned}
\scriptsize
\setlength{\tabcolsep}{2.5pt}
\begin{tabular}{@{}lrrrrrrrr@{}}
\toprule
 & \multicolumn{4}{c}{Global} & \multicolumn{4}{c}{Rolling} \\
\cmidrule(lr){2-5} \cmidrule(lr){6-9}
Method & \multicolumn{2}{c}{Fail} & \multicolumn{2}{c}{APE} & \multicolumn{2}{c}{Fail} & \multicolumn{2}{c}{APE} \\
 & D & T & D & T & D & T & D & T \\
\midrule
MOLA-LO & 1 & 2 & 0.053 & 0.058 & 1 & 2 & 0.063 & 0.053 \\
MOLA-LIO & 2 & 2 & 0.039 & 0.038 & 2 & 2 & 0.032 & 0.036 \\
DLIO & 2 & 3 & 0.135 & 0.120 & 2 & 2 & 0.118 & 0.125 \\
FAST-LIO2 & 0 & 2 & 0.146 & 0.076 & 0 & 2 & 0.152 & 0.089 \\
GLIM & 3 & 3 & 0.179 & 0.061 & 3 & 3 & 0.173 & 0.076 \\
EllipseLIO & 6 & 4 & 0.107 & 0.127 & 6 & 4 & 0.114 & 0.132 \\
KISS-ICP$^\dagger$ & 2 & 1 & 0.734 & 0.920 & 3 & 1 & 1.570 & 3.261 \\
Traj-LO$^\dagger$ & -- & -- & -- & -- & 3 & 3 & 0.137 & 0.145 \\
\bottomrule
\multicolumn{9}{@{}l}{\scriptsize $^\dagger$LiDAR only.}\\
\end{tabular}
\end{table}

\begin{table*}[t]
\centering
\caption{Paired effects on the APE: median ratio over the sequences where both runs succeed, its 95\,\% bootstrap interval (10000 resamples of the worlds, each with all its runs: the 15 worlds, or the 10 of the test split for the tuned column), in subscript the number of paired runs, and in superscript the change in the number of failures; in bold, the ratios whose interval excludes 1 (with 54 intervals, about 3 would by chance). Noise level: OS1-32, both shutters; shutter: rolling over global (R/G); sensor, rolling shutter: VLP-16 (16) over OS1-32 (32) on the \emph{noisy} level, and OS1-64 (64) over OS1-32 pooled over the \emph{noisy} and \emph{real} levels; tuned over default on the 20 test sequences, all variants. Below 1: lower error.}
\label{tab:effects}
\scriptsize
\setlength{\tabcolsep}{2.3pt}
\begin{tabular}{@{}lrrrrrrr@{}}
\toprule
 & \multicolumn{2}{c}{Noise} & \multicolumn{2}{c}{Shutter} & \multicolumn{2}{c}{Sensor} & \\
\cmidrule(lr){2-3} \cmidrule(lr){4-5} \cmidrule(lr){6-7}
Method & noisy/clean & real/noisy & R/G clean & R/G real & 16/32 & 64/32 & tuned \\
\midrule
MOLA-LO & \textbf{1.90} [1.36, 2.42]$_{58}$ & \textbf{1.31} [1.03, 1.73]$_{56}$\textsuperscript{+2} & 1.04 [1.00, 1.12]$_{29}$ & 1.01 [0.90, 1.07]$_{28}$ & \textbf{2.18} [1.73, 3.06]$_{27}$\textsuperscript{+2} & \textbf{0.76} [0.52, 0.91]$_{57}$\textsuperscript{-2} & 0.94 [0.85, 1.00]$_{169}$\textsuperscript{+7} \\
MOLA-LIO & \textbf{2.22} [1.60, 3.45]$_{58}$\textsuperscript{+1} & \textbf{1.36} [1.06, 1.81]$_{54}$\textsuperscript{+4} & 1.04 [0.99, 1.18]$_{29}$\textsuperscript{+1} & 1.02 [0.99, 1.04]$_{27}$ & \textbf{2.00} [1.40, 3.98]$_{26}$\textsuperscript{+3} & \textbf{0.75} [0.55, 0.86]$_{56}$\textsuperscript{-2} & 0.92 [0.78, 1.00]$_{169}$\textsuperscript{+4} \\
DLIO & \textbf{1.09} [1.02, 1.17]$_{56}$\textsuperscript{-1} & 1.04 [0.97, 1.40]$_{54}$\textsuperscript{+3} & 1.05 [0.99, 1.17]$_{28}$ & 1.00 [0.97, 1.06]$_{27}$ & \textbf{1.95} [1.09, 3.06]$_{25}$\textsuperscript{+3} & \textbf{0.77} [0.65, 0.85]$_{55}$\textsuperscript{-3} & 1.00 [0.98, 1.02]$_{165}$\textsuperscript{+1} \\
FAST-LIO2 & 1.06 [0.97, 1.14]$_{57}$\textsuperscript{+1} & 1.01 [0.90, 1.22]$_{56}$ & 1.01 [0.98, 1.07]$_{29}$ & 1.02 [0.96, 1.07]$_{28}$\textsuperscript{-1} & 1.04 [0.84, 1.34]$_{25}$\textsuperscript{+3} & \textbf{0.90} [0.80, 0.98]$_{57}$\textsuperscript{-2} & \textbf{0.56} [0.47, 0.91]$_{165}$\textsuperscript{+11} \\
GLIM & \textbf{1.32} [1.08, 1.49]$_{51}$\textsuperscript{+4} & \textbf{1.16} [1.02, 1.56]$_{50}$\textsuperscript{+1} & 1.02 [0.89, 1.16]$_{27}$\textsuperscript{-1} & 0.99 [0.94, 1.08]$_{25}$ & \textbf{8.61} [4.11, 18.32]$_{22}$\textsuperscript{+4} & \textbf{0.69} [0.53, 0.89]$_{51}$\textsuperscript{-3} & \textbf{0.72} [0.46, 0.87]$_{155}$\textsuperscript{+2} \\
EllipseLIO & 1.03 [0.98, 1.06]$_{45}$\textsuperscript{+2} & 1.04 [0.93, 1.22]$_{42}$\textsuperscript{+2} & 1.01 [0.99, 1.13]$_{23}$\textsuperscript{-1} & 1.03 [1.00, 1.08]$_{21}$\textsuperscript{-1} & \textbf{1.15} [1.08, 1.38]$_{15}$\textsuperscript{+7} & \textbf{0.71} [0.52, 0.78]$_{41}$\textsuperscript{-3} & 0.99 [0.96, 1.01]$_{123}$\textsuperscript{-15} \\
KISS-ICP$^\dagger$ & \textbf{1.05} [1.01, 1.10]$_{44}$ & 1.00 [0.80, 1.08]$_{44}$\textsuperscript{-5} & \textbf{1.11} [1.01, 1.37]$_{21}$\textsuperscript{+3} & \textbf{1.07} [1.03, 1.57]$_{24}$\textsuperscript{+2} & 1.14 [0.36, 1.27]$_{18}$\textsuperscript{+2} & \textbf{0.62} [0.34, 0.94]$_{42}$\textsuperscript{-1} & 1.08 [0.99, 1.27]$_{137}$\textsuperscript{-12} \\
Traj-LO$^\dagger$ & 1.03 [0.89, 1.41]$_{20}$\textsuperscript{-1} & \textbf{1.33} [1.21, 1.64]$_{20}$ & -- & -- & \textbf{1.65} [1.40, 2.46]$_{21}$\textsuperscript{-1} & 0.79 [0.54, 1.00]$_{43}$\textsuperscript{-2} & 0.92 [0.90, 1.00]$_{78}$ \\
\bottomrule
\multicolumn{8}{@{}l}{\scriptsize $^\dagger$LiDAR only; Traj-LO needs per-point times (rolling shutter only).}\\
\end{tabular}
\end{table*}

\subsection{Effect of the noise level}
Because the variants are paired, the effect of each error source is measured
on the same sequences, motions and random seeds (\cref{tab:effects}). Gaussian
range noise and the IMU noise (\emph{noisy} over \emph{clean}) double
the APE of MOLA-LO and MOLA-LIO (1.90 and 2.22) and raise GLIM's by
\SI{32}{\percent}, while DLIO, FAST-LIO2,
EllipseLIO and the LiDAR-only methods change little (median ratios
\numrange{1.03}{1.09}, with intervals above one only for DLIO and KISS-ICP);
the RTE ratios are similar. The sensitive methods are
the most accurate ones on noise-free data, where MOLA reaches
\SIrange{1.5}{4.6}{\centi\meter} of median APE against
\SIrange{11}{13}{\centi\meter} for DLIO, FAST-LIO2 and GLIM: the added noise
only shows in an estimate that is not already limited by coarser voxels or by
its own approximations. The realistic error model (\emph{real}
over \emph{noisy}) affects mostly the same methods: it raises the APE of
MOLA-LO, MOLA-LIO and Traj-LO by about a third (1.31 to 1.36) and GLIM's by
\SI{16}{\percent}, all four with intervals above one, and changes the others
by at most \SI{4}{\percent}, with intervals that include no change; the RTE
ratios are similar. Its constant noise term (\SIrange{0.5}{2}{\centi\meter})
is at most that of the \emph{noisy} level, so the difference comes from the
range-dependent noise, the beam footprint and its mixed returns, the lost weak
returns and the per-ring calibration errors. Over both shutters, it adds two
to four failures to MOLA-LO, MOLA-LIO, DLIO and EllipseLIO, one to GLIM and
none to FAST-LIO2, and removes five from KISS-ICP. The ranking by median APE
changes little: MOLA-LIO stays first with both shutters, and DLIO and
FAST-LIO2, close on both levels (0.11 to \SI{0.15}{\meter}), swap places in
opposite directions for the two shutters.

\subsection{Effect of the shutter model}
With each method's own de-skew, the median paired APE ratio of rolling over
global shutter is between 0.98 and 1.16 for every method and noise level
(\cref{tab:effects}; the \emph{noisy} level, not shown, gives the extremes), only KISS-ICP's ratios have intervals above one, and
the failures change by at most three. The cost
concentrates on the most aggressive motions. On the noise-free data, where it
is not masked by noise, KISS-ICP's constant-velocity de-skew multiplies its
APE by 2.7 on the hand-held sequences (six pairs) and by 31 on the two
legged-robot ones,
whose body pitches on stairs and slopes; the IMU-based de-skew of the other methods
keeps them within \SI{40}{\percent} on every carrier, except EllipseLIO on
the quadruped sequences (8.5, on its three completed pairs). A few ratios are well below one: on the four clean quadruped sequences, the
rolling-shutter runs of MOLA-LO have 0.16 times the APE of the global-shutter
ones (MOLA-LIO 0.56, at APEs of 1 to \SI{28}{\milli\meter}). In three of its
four global-shutter runs, MOLA-LO registers the top face of a floor slab, seen
from below before the robot steps onto it, to the underside already in its
map, and drops by the slab thickness (\SIrange{0.20}{0.25}{\meter}); DLIO, GLIM
and FAST-LIO2 have vertical errors of that size with both shutters. The
ideal, parallel faces of the simulated slabs make this easier than in real
buildings.

\subsection{Effect of the sensor}
The sensor matters more than the shutter (\cref{tab:effects}). The VLP-16
differs from the OS1-32 in more than its 16 rings: its vertical field of view
is \SI{30}{\degree} instead of \SI{45}{\degree}, it has 1808 columns instead
of 1024, and its range reaches \SI{100}{\meter} instead of \SI{60}{\meter}.
With it (both \emph{noisy}, rolling shutter), the median APE doubles for MOLA-LO, MOLA-LIO and DLIO (1.95 to
2.18), and grows by a factor of 1.65 for Traj-LO and 8.6 for GLIM, whose failures go from four to
eight; KISS-ICP and FAST-LIO2 show no clear change (1.14 and 1.04, with
intervals that include one), and EllipseLIO's APE grows by \SI{15}{\percent}
while it fails on 14 sequences instead of 7. The OS1-64
differs from the OS1-32 only in its 64 rings, and shares its error-model
draws, so it isolates the ring count: it improves every method, at both noise
levels, by \SIrange{8}{38}{\percent} of the median (the intervals of the
pooled ratio are below one for every method but Traj-LO), and removes one to
three failures from every method over the two levels.

\subsection{Map accuracy}
\Cref{tab:maps} gives the map metrics on the \emph{real} rolling-shutter
variant; the global-shutter variant gives the same ordering. With the
variant's own sweeps at the true poses, the sensor errors alone limit the map to a
median accuracy of \SI{2.1}{\centi\meter} and an F-score of 0.95 at
\SI{10}{\centi\meter}, the floor for every method. The map ranking follows
the trajectory ranking: MOLA-LIO, MOLA-LO and EllipseLIO (on its 22
successful runs) are closest to the floor, with F-scores of 0.85, 0.80 and
0.75 and a median accuracy of \SIrange{5.7}{6.0}{\centi\meter}; FAST-LIO2 and
DLIO follow, nearly tied (0.72 and 0.71), then GLIM (0.60), and KISS-ICP is
far behind (0.25).

\begin{table}[t]
\centering
\caption{Map metrics on os1\_32-real-rolling: medians over each method's successful runs of the accuracy (median distance to the reference) and completeness [m], the Chamfer distance [m] (the average of the mean accuracy and the completeness), and the F-score at 10 and \SI{20}{\centi\meter}. True poses: the variant's own sweeps at the true trajectory, the floor set by the sensor errors. Traj-LO: not computed.}
\label{tab:maps}
\footnotesize
\setlength{\tabcolsep}{3pt}
\begin{tabular}{@{}lrrrrrr@{}}
\toprule
Method & Runs & Acc. & Comp. & Chamfer & F$_{10}$ & F$_{20}$ \\
\midrule
True poses & 30 & 0.021 & 0.037 & 0.043 & 0.95 & 0.98 \\
\midrule
MOLA-LO & 28 & 0.060 & 0.054 & 0.068 & 0.80 & 0.95 \\
MOLA-LIO & 27 & 0.057 & 0.048 & 0.064 & 0.85 & 0.96 \\
DLIO & 27 & 0.075 & 0.083 & 0.089 & 0.71 & 0.91 \\
FAST-LIO2 & 29 & 0.070 & 0.081 & 0.088 & 0.72 & 0.92 \\
GLIM & 25 & 0.101 & 0.081 & 0.116 & 0.60 & 0.85 \\
EllipseLIO & 22 & 0.059 & 0.067 & 0.073 & 0.75 & 0.94 \\
KISS-ICP & 24 & 0.311 & 0.286 & 0.350 & 0.25 & 0.46 \\
\bottomrule
\end{tabular}
\end{table}

\subsection{Map extent}
\label{sec:map-extent}
For MOLA, a \SI{1}{\kilo\meter} map instead of the \SI{100}{\meter} default
changes the APE by less than \SI{10}{\percent} on 23 to 25 of the 27 or 28
sequences each configuration completes on each \emph{real} variant (median
ratios of the APE and the RTE of 1.00); on the loops that return to places
left behind, re-registering against the old map lowers the APE by 14 to
\SI{26}{\percent} in six of the eight runs on the two parking-lot sequences
and raises the RTE, as expected of an implicit loop closure. FAST-LIO2 is more
sensitive: with a \SI{200}{\meter} box instead of \SI{1}{\kilo\meter}, its APE
changes by more than \SI{10}{\percent} on 9 of 28 and 12 of 29 runs, in both
directions (\texttt{campus/long\_tour}, rolling shutter: 2.4 to
\SI{5.8}{\meter}; \texttt{open\_field/lap}: 5.5 to \SI{1.9}{\meter}), and no
run fails with one box and completes with the other. APE comparisons between
methods with different map extents should therefore be read together with
the RTE (\cref{tab:main_summary}).

\subsection{Runtime}
\label{sec:runtime}
Every run of the synthetic variants was repeated on a desktop CPU
(Ryzen~7 2700X at about \SI{4}{\giga\hertz}), single-threaded and pinned to one
physical core whose second hardware thread was kept idle. Every run ended with
the same outcome as in the accuracy runs, and the 1994 successful runs with
the same APE to all printed digits. \Cref{tab:runtime} gives the real-time
factors. With their defaults on the OS1-32, every LiDAR-inertial method runs
faster than real time on one core, from 1.7 times (MOLA-LIO) to 3.2 times
(FAST-LIO2); KISS-ICP barely keeps up (1.1) and Traj-LO does not (0.75).
Doubling the rings (OS1-64) slows every method: by 8 and \SI{10}{\percent}
MOLA-LIO and MOLA-LO, by \SI{23}{\percent} GLIM, by 35 to \SI{43}{\percent}
FAST-LIO2, Traj-LO and DLIO, and by more than half KISS-ICP and EllipseLIO.
Run for run on the test sequences that both configurations complete, the
tuned ones take about 1.9 times longer than the defaults for FAST-LIO2 and
Traj-LO, 3.4 times for KISS-ICP, about 1.4 times for MOLA-LO and MOLA-LIO,
and up to \SI{11}{\percent} longer for the others.

\begin{table}[t]
\centering
\caption{Real-time factor on one core of a desktop CPU: sequence duration over the wall time of the run, median over the successful runs of the \emph{noisy} rolling-shutter variants. Defaults on the 30 sequences, and the tuned configuration (T) on the 20 test sequences.}
\label{tab:runtime}
\footnotesize
\setlength{\tabcolsep}{4pt}
\begin{tabular}{@{}lrrrr@{}}
\toprule
Method & VLP-16 & OS1-32 & OS1-64 & OS1-32 T \\
\midrule
MOLA-LO & 2.06 & 1.92 & 1.73 & 1.38 \\
MOLA-LIO & 1.79 & 1.69 & 1.57 & 1.20 \\
DLIO & 2.79 & 2.39 & 1.35 & 2.07 \\
FAST-LIO2 & 3.38 & 3.17 & 2.06 & 1.19 \\
GLIM & 2.15 & 2.77 & 2.12 & 2.42 \\
EllipseLIO & 4.04 & 2.45 & 1.02 & 1.96 \\
KISS-ICP$^\dagger$ & 1.63 & 1.10 & 0.50 & 0.22 \\
Traj-LO$^\dagger$ & 0.62 & 0.75 & 0.44 & 0.29 \\
\bottomrule
\multicolumn{5}{@{}l}{\scriptsize $^\dagger$LiDAR only.}\\
\end{tabular}
\end{table}

\subsection{Scene difficulty}
\label{sec:difficulty}
To report results by difficulty without letting any single method define it,
each sequence is scored by the median, over the seven methods that run on
global shutter, with their defaults, of the APE as a percentage of the path length on the clean global-shutter
variant, where only the scene geometry and the motion make a sequence hard
(a failure counts as \SI{10}{\percent}). The tertiles of this score give
three classes of ten sequences. \emph{Easy}: urban street, greenhouse,
warehouse and parking lot (all two sequences each), orchard rows and the
parking garage drive. \emph{Hard}: tunnel, open field, campus and rough
terrain (two each), and the quadruped on the apartment stairs and in the stairwell.
The rest are \emph{moderate}. Since the classes come from the methods being
compared, they describe the difficulty for current methods rather than a
property of the scene alone (\cref{sec:observability} measures that
property). The results per class are released with the evaluation code.

Every method completes the easy and moderate classes on the \emph{real}
variants, except EllipseLIO (2 failures of 20 in the moderate class). The
hard class separates them (failures of 20 runs, both shutters): FAST-LIO2 3,
MOLA-LO 4, MOLA-LIO and DLIO 6, KISS-ICP and GLIM 10, and EllipseLIO 15
(Traj-LO, rolling shutter only: 4 of 7). Two scenes defeat most methods. In
the open field, the ground constrains height, roll and pitch, but the
horizontal position and the yaw rest on a handful of distant poles and trees:
on the \emph{real} OS1-32 variants, only MOLA-LO, KISS-ICP and FAST-LIO2
complete the lap, and only KISS-ICP and FAST-LIO2 the zigzag, with APEs of 5.5
to \SI{58}{\meter} (with the denser OS1-64, MOLA-LO, MOLA-LIO, DLIO and
FAST-LIO2 complete both, with 2 to \SI{24}{\meter}). In the tunnel, the
repeated cross-section leaves the motion along the axis to a few fixtures
(\cref{tab:observability}). The slow
inspection run defeats every method on the \emph{real} OS1-32 variants but
EllipseLIO with rolling shutter (\SI{33}{\meter}), whereas Traj-LO completes
it on its five other variants; the fast car run is completed with both
shutters only by the MOLA configurations and DLIO, with an APE of 2 to
\SI{11}{\meter}. The best runs on the \emph{real} OS1-32 variants (2.3 and
\SI{33}{\meter} in the tunnel, 5.5 and \SI{8.9}{\meter} in the open field) are
10 to 140 times the drift of an ideal estimator (\cref{sec:observability}).
The other failures are not nested. Every method but
EllipseLIO (and Traj-LO, stopped at the memory limit) completes the car on the campus ring road
(\SI{8}{\meter\per\second}), but MOLA and KISS-ICP drift there to 32 to
\SI{69}{\meter} of APE, where FAST-LIO2 and GLIM keep 3 to \SI{6}{\meter}
(DLIO \SI{2}{\meter} with global shutter and \SI{158}{\meter} with rolling
shutter); only the MOLA configurations and DLIO complete the rough-terrain
loop with an APE under \SI{4}{\meter} (FAST-LIO2: 37 to \SI{38}{\meter}); and
EllipseLIO fails in the stairwell with the quadruped, which every other method
completes with global shutter.

\section{Discussion and Limitations}
\label{sec:discussion}
\begin{figure}[t]
\centering
\includegraphics[width=\columnwidth]{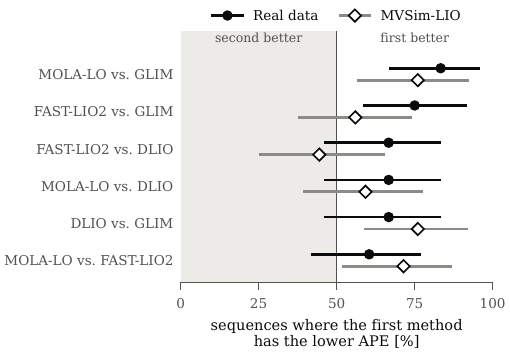}
\caption{\bench{} against real data, two methods at a time: for each pair,
the share of sequences on which the first method has the lower APE (at the
millimeter; a tie counts one half), among those both complete, with its
95\,\% interval. Real data: 24 Oxford Spires, GrandTour and
FoMo sequences; \bench{}: os1\_32-real-rolling. The first method
of each pair is the one more often better on real data, so a \bench{} marker
in the shaded half picks the other method.}
\label{fig:sim2real}
\end{figure}

\paragraph{Do the rankings transfer to real data?}
We ran four of the methods, with the same builds, on 24 real sequences: 13 of
Oxford Spires \citep{tao2025spires}, whose reference registers the scans to a
TLS map, 7 GrandTour \citep{tuna2026grandtour} missions with an RTK GNSS/INS
reference, and 4 FoMo \citep{boxan2026fomo} deployments with GNSS positions
only. Each run is scored by its APE after SE(3) alignment to its own
reference, and none fails. \Cref{fig:sim2real} compares the methods two at a
time: for each pair, the share of sequences on which one has the lower APE, on
real data and on \bench{}. The benchmark picks the same winner as the real
data for five of the six pairs. The exception is FAST-LIO2 against DLIO:
FAST-LIO2 is the more accurate on \SI{67}{\percent} of the real sequences but
on \SI{44}{\percent} of the simulated ones, and both intervals include an even
split. With 24 to 28 sequences per pair, the intervals are wide, so the
agreement shows which method each benchmark prefers, not by how much.

\paragraph{Limitations}
Exact ground truth comes at the price of a simplified world. The ray tracer
supports analytic primitives only: vegetation is made of trunks, branches and
clusters of spheres, and cars of boxes and wheels. There is no glass, no
intensity (which rules out intensity-aided methods \citep{pfreundschuh2024coinlio}),
no multi-echo, no cameras, no weather, and no dynamic objects, all of which
earlier simulated datasets offer in part \citep{han2023carlaloc,patel2025tartanground}.
The worlds are dominated by planes and simple solids, which may favor
plane-based registration. The IMU has white noise and
bias random walk but no initial bias, scale factor or axis misalignment; the
extrinsics are exact, all sensors share one clock, and the legged gaits have
no foot impacts or vibration. The parameters of the realistic error model are
drawn once per sequence, so their effect is not separated from that of the
scene. Compared with real sensors (\cref{sec:validation}), it is noisier
than most real units at extreme grazing incidence (85--\SI{90}{\degree}), loses no
returns below grazing incidence at short range, where real sensors lose 0.1 to
\SI{3}{\percent}, and its per-ring biases could not be checked against the
reference mesh. Its beam is
sampled by seven rays, so objects that fall between them do not affect the
return. The OS1 rigs stop at \SI{60}{\meter}, short of the sensors' specified
range, which limits the structure visible in the open field. Each of
these effects can be changed in the generator and rendered as a further
paired variant.
Results on \bench{} complement evaluation on real data by isolating how an
estimator responds to each error source.

\section{Conclusions}
\label{sec:conclusions}
\bench{} separates the factors that real benchmarks confound. Its paired
variants show that the realistic LiDAR error model degrades mostly the
methods that the \emph{noisy} level (Gaussian range and IMU noise) degrades
(MOLA-LO, MOLA-LIO and GLIM, plus Traj-LO) without changing the ranking much,
that each method's de-skew absorbs most of the rolling shutter, and that the
sensor matters more than both for some methods (GLIM) and hardly at all for
others (FAST-LIO2). Most methods fail in the open field and the tunnel,
although a method-independent analysis finds both observable: there the
motion rests on a weak constraint or on a few dozen points per sweep. The
error model's ranges bracket two real LiDARs in noise (except at extreme
grazing incidence, where it is noisier than most units) and in mixed returns
at depth edges, and it keeps returning at grazing incidence as they do,
although it loses none of the 0.1 to \SI{3}{\percent} of returns they lose;
three more Ouster units of public datasets sit at the low end of its noise
ranges. On 24 real sequences, \bench{} picks the more accurate of two methods
as the real data does for five of the six pairs compared. Of the methods, MOLA-LIO has
the lowest median APE with its default parameters (on the sequences that
both complete, \SI{7}{\percent} below MOLA-LO's, and not distinguishable from
DLIO's), FAST-LIO2 fails on the fewest sequences, and which method completes
the most runs depends on the accuracy demanded of a run; tuning on a training
split lowers the APE of FAST-LIO2 and GLIM by \SI{44}{\percent} and
\SI{28}{\percent}, but removes failures only for EllipseLIO and KISS-ICP. The
data, generator and evaluation tools are openly available (see the data
availability statement).

\begin{dci}
J.L.~Blanco-Claraco develops MOLA and MVSim, which are used and evaluated in
this work. The other authors declared no potential conflicts of interest with
respect to the research, authorship, and/or publication of this article.
\end{dci}

\begin{funding}
The authors received no financial support for the research, authorship, and/or
publication of this article.
\end{funding}

\begin{dataavail}
The dataset is openly available at
\url{https://huggingface.co/datasets/MOLAorg/MVSim-LIO} under the CC BY 4.0
license; this article uses its release tagged \texttt{v2.1}, rendered with
MVSim commit \texttt{8a45b9e1}. The worlds,
trajectories, generator, ground truth tools, method configurations and
evaluation code are available at
\url{https://github.com/MOLAorg/mvsim_lio_benchmark}
under the BSD 3-Clause license.
\end{dataavail}

\bibliographystyle{SageH}
\bibliography{refs}

\begin{thebibliography}{59}
\providecommand{\natexlab}[1]{#1}
\providecommand{\url}[1]{\texttt{#1}}
\providecommand{\urlprefix}{URL }
\expandafter\ifx\csname urlstyle\endcsname\relax
  \providecommand{\doi}[1]{DOI:\discretionary{}{}{}#1}\else
  \providecommand{\doi}{DOI:\discretionary{}{}{}\begingroup
  \urlstyle{rm}\Url}\fi

\bibitem[{Behley et~al.(2019)Behley, Garbade, Milioto, Quenzel, Behnke,
  Stachniss and Gall}]{behley2019semantickitti}
Behley J, Garbade M, Milioto A, Quenzel J, Behnke S, Stachniss C and Gall J
  (2019) {SemanticKITTI}: A dataset for semantic scene understanding of {LiDAR}
  sequences.
\newblock In: \emph{IEEE/CVF International Conference on Computer Vision
  (ICCV)}. pp. 9296--9306.
\newblock \doi{10.1109/ICCV.2019.00939}.

\bibitem[{Blanco-Claraco(2025)}]{blanco2025mola}
Blanco-Claraco JL (2025) A flexible framework for accurate {LiDAR} odometry,
  map manipulation, and localization.
\newblock \emph{The International Journal of Robotics Research} 44(9):
  1553--1599.
\newblock \doi{10.1177/02783649251316881}.

\bibitem[{Blanco-Claraco et~al.(2023)Blanco-Claraco, Tymchenko,
  Ma{\~n}as-Alvarez, Ca{\~n}adas-Ar{\'a}nega, L{\'o}pez-G{\'a}zquez and
  Moreno}]{blanco2023mvsim}
Blanco-Claraco JL, Tymchenko B, Ma{\~n}as-Alvarez FJ, Ca{\~n}adas-Ar{\'a}nega
  F, L{\'o}pez-G{\'a}zquez {\'A} and Moreno JC (2023) {MultiVehicle Simulator
  (MVSim)}: Lightweight dynamics simulator for multiagents and mobile robotics
  research.
\newblock \emph{SoftwareX} 23: 101443.
\newblock \doi{10.1016/j.softx.2023.101443}.

\bibitem[{Border and Chli(2026)}]{ellipselio}
Border R and Chli M (2026) {EllipseLIO}: Adaptive {LiDAR} inertial odometry
  with an ellipsoid representation.
\newblock \emph{IEEE Robotics and Automation Letters} 11(10): 11474--11481.
\newblock \doi{10.1109/LRA.2026.3726372}.

\bibitem[{Boxan et~al.(2026)Boxan, Jeanson, Krawciw, Daum, Qiao, Lilge, Barfoot
  and Pomerleau}]{boxan2026fomo}
Boxan M, Jeanson G, Krawciw A, Daum E, Qiao X, Lilge S, Barfoot TD and
  Pomerleau F (2026) {FoMo}: A multi-season dataset for robot navigation in
  {For{\^e}t Montmorency}.
\newblock \emph{The International Journal of Robotics Research}
  \doi{10.1177/02783649261491481}.
\newblock OnlineFirst.

\bibitem[{Brizi et~al.(2024)Brizi, Giacomini, Di~Giammarino, Ferrari, Salem,
  De~Rebotti and Grisetti}]{brizi2024vbr}
Brizi L, Giacomini E, Di~Giammarino L, Ferrari S, Salem O, De~Rebotti L and
  Grisetti G (2024) {VBR}: A vision benchmark in {Rome}.
\newblock In: \emph{IEEE International Conference on Robotics and Automation
  (ICRA)}. pp. 15868--15874.
\newblock \doi{10.1109/ICRA57147.2024.10611395}.

\bibitem[{Censi(2007)}]{censi2007achievable}
Censi A (2007) On achievable accuracy for range-finder localization.
\newblock In: \emph{IEEE International Conference on Robotics and Automation
  (ICRA)}. pp. 4170--4175.
\newblock \doi{10.1109/ROBOT.2007.364120}.

\bibitem[{Chen et~al.(2023)Chen, Nemiroff and Lopez}]{chen2023dlio}
Chen K, Nemiroff R and Lopez BT (2023) Direct {LiDAR}-inertial odometry:
  Lightweight {LIO} with continuous-time motion correction.
\newblock In: \emph{IEEE International Conference on Robotics and Automation
  (ICRA)}. pp. 3983--3989.
\newblock \doi{10.1109/ICRA48891.2023.10160508}.

\bibitem[{Chen et~al.(2026)Chen, Qi, Feng, Zhuang, Chen, Hu, Wu, Peng and
  Lu}]{chen2025geode}
Chen Z, Qi Y, Feng D, Zhuang X, Chen H, Hu X, Wu J, Peng K and Lu P (2026)
  Heterogeneous {LiDAR} dataset for benchmarking robust localization in diverse
  degenerate scenarios.
\newblock \emph{The International Journal of Robotics Research} 45(1): 6--22.
\newblock \doi{10.1177/02783649251344967}.

\bibitem[{Deschaud(2021)}]{deschaud2021kitticarla}
Deschaud JE (2021) {KITTI-CARLA}: a {KITTI}-like dataset generated by {CARLA}
  simulator.
\newblock arXiv:2109.00892.

\bibitem[{Deschaud et~al.(2021)Deschaud, Duque, Richa, Velasco-Forero,
  Marcotegui and Goulette}]{deschaud2021pariscarla}
Deschaud JE, Duque D, Richa JP, Velasco-Forero S, Marcotegui B and Goulette F
  (2021) {Paris-CARLA-3D}: A real and synthetic outdoor point cloud dataset for
  challenging tasks in {3D} mapping.
\newblock \emph{Remote Sensing} 13(22): 4713.
\newblock \doi{10.3390/rs13224713}.

\bibitem[{Dosovitskiy et~al.(2017)Dosovitskiy, Ros, Codevilla, Lopez and
  Koltun}]{dosovitskiy2017carla}
Dosovitskiy A, Ros G, Codevilla F, Lopez A and Koltun V (2017) {CARLA}: An open
  urban driving simulator.
\newblock In: \emph{Proceedings of the 1st Annual Conference on Robot Learning
  (CoRL)}, \emph{Proceedings of Machine Learning Research}, volume~78. pp.
  1--16.
\newblock
  \urlprefix\url{https://proceedings.mlr.press/v78/dosovitskiy17a.html}.

\bibitem[{Forster et~al.(2017)Forster, Carlone, Dellaert and
  Scaramuzza}]{forster2017preintegration}
Forster C, Carlone L, Dellaert F and Scaramuzza D (2017) On-manifold
  preintegration for real-time visual--inertial odometry.
\newblock \emph{IEEE Transactions on Robotics} 33(1): 1--21.
\newblock \doi{10.1109/TRO.2016.2597321}.

\bibitem[{Geiger et~al.(2013)Geiger, Lenz, Stiller and
  Urtasun}]{geiger2013kitti}
Geiger A, Lenz P, Stiller C and Urtasun R (2013) Vision meets robotics: The
  {KITTI} dataset.
\newblock \emph{The International Journal of Robotics Research} 32(11):
  1231--1237.
\newblock \doi{10.1177/0278364913491297}.

\bibitem[{Geiger et~al.(2012)Geiger, Lenz and Urtasun}]{geiger2012kitti}
Geiger A, Lenz P and Urtasun R (2012) Are we ready for autonomous driving? {The
  KITTI} vision benchmark suite.
\newblock In: \emph{IEEE Conference on Computer Vision and Pattern Recognition
  (CVPR)}. pp. 3354--3361.
\newblock \doi{10.1109/CVPR.2012.6248074}.

\bibitem[{Glennie and Lichti(2010)}]{glennie2010hdl64}
Glennie C and Lichti DD (2010) Static calibration and analysis of the {Velodyne
  HDL-64E S2} for high accuracy mobile scanning.
\newblock \emph{Remote Sensing} 2(6): 1610--1624.
\newblock \doi{10.3390/rs2061610}.

\bibitem[{Han et~al.(2023)Han, Liu, Sun, Li, Sun, Yuan and
  Ang}]{han2023carlaloc}
Han Y, Liu Z, Sun S, Li D, Sun J, Yuan C and Ang MH (2023) {CARLA-Loc}:
  Synthetic {SLAM} dataset with full-stack sensor setup in challenging weather
  and dynamic environments.
\newblock arXiv:2309.08909.

\bibitem[{Handa et~al.(2014)Handa, Whelan, McDonald and
  Davison}]{handa2014iclnuim}
Handa A, Whelan T, McDonald J and Davison AJ (2014) A benchmark for {RGB-D}
  visual odometry, {3D} reconstruction and {SLAM}.
\newblock In: \emph{IEEE International Conference on Robotics and Automation
  (ICRA)}. pp. 1524--1531.
\newblock \doi{10.1109/ICRA.2014.6907054}.

\bibitem[{Hebert and Krotkov(1992)}]{hebert1992mixed}
Hebert M and Krotkov E (1992) {3D} measurements from imaging laser radars: How
  good are they?
\newblock \emph{Image and Vision Computing} 10(3): 170--178.
\newblock \doi{10.1016/0262-8856(92)90068-E}.

\bibitem[{Helmberger et~al.(2022)Helmberger, Morin, Berner, Kumar, Cioffi and
  Scaramuzza}]{helmberger2022hilti}
Helmberger M, Morin K, Berner B, Kumar N, Cioffi G and Scaramuzza D (2022) The
  {Hilti SLAM} challenge dataset.
\newblock \emph{IEEE Robotics and Automation Letters} 7(3): 7518--7525.
\newblock \doi{10.1109/LRA.2022.3183759}.

\bibitem[{Jung et~al.(2024)Jung, Yang, Lee, Gil, Kim and Kim}]{jung2024helipr}
Jung M, Yang W, Lee D, Gil H, Kim G and Kim A (2024) {HeLiPR}: Heterogeneous
  {LiDAR} dataset for inter-{LiDAR} place recognition under spatiotemporal
  variations.
\newblock \emph{The International Journal of Robotics Research} 43(12):
  1867--1883.
\newblock \doi{10.1177/02783649241242136}.

\bibitem[{Kim et~al.(2020)Kim, Park, Cho, Jeong and Kim}]{kim2020mulran}
Kim G, Park YS, Cho Y, Jeong J and Kim A (2020) {MulRan}: Multimodal range
  dataset for urban place recognition.
\newblock In: \emph{IEEE International Conference on Robotics and Automation
  (ICRA)}. pp. 6246--6253.
\newblock \doi{10.1109/ICRA40945.2020.9197298}.

\bibitem[{Koenig and Howard(2004)}]{koenig2004gazebo}
Koenig N and Howard A (2004) Design and use paradigms for {Gazebo}, an
  open-source multi-robot simulator.
\newblock In: \emph{IEEE/RSJ International Conference on Intelligent Robots and
  Systems (IROS)}, volume~3. pp. 2149--2154.
\newblock \doi{10.1109/IROS.2004.1389727}.

\bibitem[{Koide et~al.(2024)Koide, Yokozuka, Oishi and Banno}]{koide2024glim}
Koide K, Yokozuka M, Oishi S and Banno A (2024) {GLIM}: {3D} range-inertial
  localization and mapping with {GPU}-accelerated scan matching factors.
\newblock \emph{Robotics and Autonomous Systems} 179: 104750.
\newblock \doi{10.1016/j.robot.2024.104750}.

\bibitem[{Kong et~al.(2023)Kong, Liu, Tang, Lin, Ren, Cai, Zhu, Chen and
  Zhang}]{kong2023marsim}
Kong F, Liu X, Tang B, Lin J, Ren Y, Cai Y, Zhu F, Chen N and Zhang F (2023)
  {MARSIM}: A light-weight point-realistic simulator for {LiDAR}-based {UAVs}.
\newblock \emph{IEEE Robotics and Automation Letters} 8(5): 2954--2961.
\newblock \doi{10.1109/LRA.2023.3264163}.

\bibitem[{Laconte et~al.(2019)Laconte, Desch{\^e}nes, Labussi{\`e}re and
  Pomerleau}]{laconte2019bias}
Laconte J, Desch{\^e}nes SP, Labussi{\`e}re M and Pomerleau F (2019) Lidar
  measurement bias estimation via return waveform modelling in a context of
  {3D} mapping.
\newblock In: \emph{IEEE International Conference on Robotics and Automation
  (ICRA)}. pp. 8100--8106.
\newblock \doi{10.1109/ICRA.2019.8793671}.

\bibitem[{Li et~al.(2022)Li, Yu, Pe{\~n}a~Queralta and
  Westerlund}]{li2022tiers}
Li Q, Yu X, Pe{\~n}a~Queralta J and Westerlund T (2022) Multi-modal lidar
  dataset for benchmarking general-purpose localization and mapping algorithms.
\newblock In: \emph{IEEE/RSJ International Conference on Intelligent Robots and
  Systems (IROS)}. pp. 3837--3844.
\newblock \doi{10.1109/IROS47612.2022.9981078}.

\bibitem[{Liu et~al.(2024)Liu, Fu, Qin, Xu, Xu, Chen, Goossens, Sun, Yu, Liu,
  Chen, Tao and Zhao}]{liu2024botanic}
Liu Y, Fu Y, Qin M, Xu Y, Xu B, Chen F, Goossens B, Sun PZH, Yu H, Liu C, Chen
  L, Tao W and Zhao H (2024) {BotanicGarden}: A high-quality dataset for robot
  navigation in unstructured natural environments.
\newblock \emph{IEEE Robotics and Automation Letters} 9(3): 2798--2805.
\newblock \doi{10.1109/LRA.2024.3359548}.

\bibitem[{Manivasagam et~al.(2020)Manivasagam, Wang, Wong, Zeng, Sazanovich,
  Tan, Yang, Ma and Urtasun}]{manivasagam2020lidarsim}
Manivasagam S, Wang S, Wong K, Zeng W, Sazanovich M, Tan S, Yang B, Ma WC and
  Urtasun R (2020) {LiDARsim}: Realistic {LiDAR} simulation by leveraging the
  real world.
\newblock In: \emph{IEEE/CVF Conference on Computer Vision and Pattern
  Recognition (CVPR)}. pp. 11164--11173.
\newblock \doi{10.1109/CVPR42600.2020.01118}.

\bibitem[{Nair et~al.(2024)Nair, Kindle, Levchev and
  Scaramuzza}]{nair2024hilti2023}
Nair AD, Kindle J, Levchev P and Scaramuzza D (2024) {Hilti SLAM} challenge
  2023: Benchmarking single + multi-session {SLAM} across sensor constellations
  in construction.
\newblock \emph{IEEE Robotics and Automation Letters} 9(8): 7286--7293.
\newblock \doi{10.1109/LRA.2024.3421791}.

\bibitem[{Nguyen et~al.(2022)Nguyen, Yuan, Cao, Lyu, Nguyen and
  Xie}]{nguyen2022ntuviral}
Nguyen TM, Yuan S, Cao M, Lyu Y, Nguyen TH and Xie L (2022) {NTU VIRAL}: A
  visual-inertial-ranging-lidar dataset, from an aerial vehicle viewpoint.
\newblock \emph{The International Journal of Robotics Research} 41(3):
  270--280.
\newblock \doi{10.1177/02783649211052312}.

\bibitem[{Nguyen et~al.(2024)Nguyen, Yuan, Nguyen, Yin, Cao, Xie, Wozniak,
  Jensfelt, Thiel, Ziegenbein and Blunder}]{nguyen2024mcd}
Nguyen TM, Yuan S, Nguyen TH, Yin P, Cao H, Xie L, Wozniak M, Jensfelt P, Thiel
  M, Ziegenbein J and Blunder N (2024) {MCD}: Diverse large-scale multi-campus
  dataset for robot perception.
\newblock In: \emph{IEEE/CVF Conference on Computer Vision and Pattern
  Recognition (CVPR)}. pp. 22304--22313.
\newblock \doi{10.1109/CVPR52733.2024.02105}.

\bibitem[{Patel et~al.(2025)Patel, Yang, Qiu, Cadena, Scherer, Hutter and
  Wang}]{patel2025tartanground}
Patel M, Yang F, Qiu Y, Cadena C, Scherer S, Hutter M and Wang W (2025)
  {TartanGround}: A large-scale dataset for ground robot perception and
  navigation.
\newblock In: \emph{IEEE/RSJ International Conference on Intelligent Robots and
  Systems (IROS)}. pp. 20524--20531.
\newblock \doi{10.1109/IROS60139.2025.11246002}.

\bibitem[{Pfreundschuh et~al.(2024)Pfreundschuh, Oleynikova, Cadena, Siegwart
  and Andersson}]{pfreundschuh2024coinlio}
Pfreundschuh P, Oleynikova H, Cadena C, Siegwart R and Andersson O (2024)
  {COIN-LIO}: Complementary intensity-augmented {LiDAR} inertial odometry.
\newblock In: \emph{IEEE International Conference on Robotics and Automation
  (ICRA)}. pp. 1730--1737.
\newblock \doi{10.1109/ICRA57147.2024.10610938}.

\bibitem[{Ramezani et~al.(2020)Ramezani, Wang, Camurri, Wisth, Mattamala and
  Fallon}]{ramezani2020newer}
Ramezani M, Wang Y, Camurri M, Wisth D, Mattamala M and Fallon M (2020) The
  {Newer College} dataset: Handheld {LiDAR}, inertial and vision with ground
  truth.
\newblock In: \emph{IEEE/RSJ International Conference on Intelligent Robots and
  Systems (IROS)}. pp. 4353--4360.
\newblock \doi{10.1109/IROS45743.2020.9340849}.

\bibitem[{Segal et~al.(2009)Segal, Haehnel and Thrun}]{segal2009gicp}
Segal A, Haehnel D and Thrun S (2009) {Generalized-ICP}.
\newblock In: \emph{Robotics: Science and Systems V}. Seattle, WA, USA.
\newblock \doi{10.15607/RSS.2009.V.021}.

\bibitem[{Shah et~al.(2018)Shah, Dey, Lovett and Kapoor}]{shah2017airsim}
Shah S, Dey D, Lovett C and Kapoor A (2018) {AirSim}: High-fidelity visual and
  physical simulation for autonomous vehicles.
\newblock In: Hutter M and Siegwart R (eds.) \emph{Field and Service Robotics},
  \emph{Springer Proceedings in Advanced Robotics}, volume~5. Springer, pp.
  621--635.
\newblock \doi{10.1007/978-3-319-67361-5_40}.

\bibitem[{Song et~al.(2025)Song, Lee, Oh, Jung and Kim}]{song2025cns}
Song H, Lee D, Oh S, Jung M and Kim A (2025) The city that never settles:
  Simulation-based {LiDAR} dataset for long-term place recognition under
  extreme structural changes.
\newblock arXiv:2505.05076.

\bibitem[{Soudarissanane et~al.(2011)Soudarissanane, Lindenbergh, Menenti and
  Teunissen}]{soudarissanane2011scanning}
Soudarissanane S, Lindenbergh R, Menenti M and Teunissen P (2011) Scanning
  geometry: Influencing factor on the quality of terrestrial laser scanning
  points.
\newblock \emph{ISPRS Journal of Photogrammetry and Remote Sensing} 66(4):
  389--399.
\newblock \doi{10.1016/j.isprsjprs.2011.01.005}.

\bibitem[{Sturm et~al.(2012)Sturm, Engelhard, Endres, Burgard and
  Cremers}]{sturm2012tum}
Sturm J, Engelhard N, Endres F, Burgard W and Cremers D (2012) A benchmark for
  the evaluation of {RGB-D} {SLAM} systems.
\newblock In: \emph{IEEE/RSJ International Conference on Intelligent Robots and
  Systems (IROS)}. pp. 573--580.
\newblock \doi{10.1109/IROS.2012.6385773}.

\bibitem[{Sun et~al.(2023)Sun, Ding, Qiu, Yoshiyasu and
  Kanehiro}]{sun2023transfusionodom}
Sun L, Ding G, Qiu Y, Yoshiyasu Y and Kanehiro F (2023) {TransFusionOdom}:
  Transformer-based {LiDAR}-inertial fusion odometry estimation.
\newblock \emph{IEEE Sensors Journal} 23(18): 22064--22079.
\newblock \doi{10.1109/JSEN.2023.3302401}.

\bibitem[{Tao et~al.(2026)Tao, Mu{\~n}oz-Ba{\~n}{\'o}n, Zhang, Wang, Fu and
  Fallon}]{tao2025spires}
Tao Y, Mu{\~n}oz-Ba{\~n}{\'o}n M{\'A}, Zhang L, Wang J, Fu LFT and Fallon M
  (2026) The {Oxford Spires} dataset: Benchmarking large-scale {LiDAR}-visual
  localisation, reconstruction and radiance field methods.
\newblock \emph{The International Journal of Robotics Research} 45(6):
  839--857.
\newblock \doi{10.1177/02783649251369905}.

\bibitem[{Tuley et~al.(2005)Tuley, Vandapel and Hebert}]{tuley2005artifacts}
Tuley J, Vandapel N and Hebert M (2005) Analysis and removal of artifacts in
  {3-D} {LADAR} data.
\newblock In: \emph{IEEE International Conference on Robotics and Automation
  (ICRA)}. pp. 2203--2210.
\newblock \doi{10.1109/ROBOT.2005.1570440}.

\bibitem[{Tuna et~al.(2026)Tuna, Frey, Fu, Patterson, Xu, Fallon, Cadena and
  Hutter}]{tuna2026grandtour}
Tuna T, Frey J, Fu LFT, Patterson K, Xu T, Fallon M, Cadena C and Hutter M
  (2026) {GrandTour}: A legged robotics dataset in the wild for multi-modal
  perception and state estimation.
\newblock \emph{The International Journal of Robotics Research}
  \doi{10.1177/02783649261480737}.
\newblock OnlineFirst.

\bibitem[{Tuna et~al.(2024)Tuna, Nubert, Nava, Khattak and
  Hutter}]{tuna2024xicp}
Tuna T, Nubert J, Nava Y, Khattak S and Hutter M (2024) {X-ICP}:
  Localizability-aware {LiDAR} registration for robust localization in extreme
  environments.
\newblock \emph{IEEE Transactions on Robotics} 40: 452--471.
\newblock \doi{10.1109/TRO.2023.3335691}.

\bibitem[{Umeyama(1991)}]{umeyama1991}
Umeyama S (1991) Least-squares estimation of transformation parameters between
  two point patterns.
\newblock \emph{IEEE Transactions on Pattern Analysis and Machine Intelligence}
  13(4): 376--380.
\newblock \doi{10.1109/34.88573}.

\bibitem[{Vizzo et~al.(2023)Vizzo, Guadagnino, Mersch, Wiesmann, Behley and
  Stachniss}]{vizzo2023kissicp}
Vizzo I, Guadagnino T, Mersch B, Wiesmann L, Behley J and Stachniss C (2023)
  {KISS-ICP}: In defense of point-to-point {ICP} -- simple, accurate, and
  robust registration if done the right way.
\newblock \emph{IEEE Robotics and Automation Letters} 8(2): 1029--1036.
\newblock \doi{10.1109/LRA.2023.3236571}.

\bibitem[{Wang et~al.(2020)Wang, Zhu, Wang, Hu, Qiu, Wang, Hu, Kapoor and
  Scherer}]{wang2020tartanair}
Wang W, Zhu D, Wang X, Hu Y, Qiu Y, Wang C, Hu Y, Kapoor A and Scherer S (2020)
  {TartanAir}: A dataset to push the limits of visual {SLAM}.
\newblock In: \emph{IEEE/RSJ International Conference on Intelligent Robots and
  Systems (IROS)}. pp. 4909--4916.
\newblock \doi{10.1109/IROS45743.2020.9341801}.

\bibitem[{Wei et~al.(2025)Wei, Jiao, Hu, Yu, Xie, Wu, Zhu, Liu, Wang and
  Liu}]{wei2025fusionportablev2}
Wei H, Jiao J, Hu X, Yu J, Xie X, Wu J, Zhu Y, Liu Y, Wang L and Liu M (2025)
  {FusionPortableV2}: A unified multi-sensor dataset for generalized {SLAM}
  across diverse platforms and scalable environments.
\newblock \emph{The International Journal of Robotics Research} 44(7):
  1093--1116.
\newblock \doi{10.1177/02783649241303525}.

\bibitem[{Winiwarter et~al.(2022)Winiwarter, Esmor{\'i}s~Pena, Weiser, Anders,
  Mart{\'i}nez~S{\'a}nchez, Searle and H{\"o}fle}]{winiwarter2022helios}
Winiwarter L, Esmor{\'i}s~Pena AM, Weiser H, Anders K, Mart{\'i}nez~S{\'a}nchez
  J, Searle M and H{\"o}fle B (2022) Virtual laser scanning with {HELIOS++}: A
  novel take on ray tracing-based simulation of topographic full-waveform {3D}
  laser scanning.
\newblock \emph{Remote Sensing of Environment} 269: 112772.
\newblock \doi{10.1016/j.rse.2021.112772}.

\bibitem[{Xiao et~al.(2022)Xiao, Huang, Guan, Zhan and Lu}]{xiao2022synlidar}
Xiao A, Huang J, Guan D, Zhan F and Lu S (2022) Transfer learning from
  synthetic to real {LiDAR} point cloud for semantic segmentation.
\newblock In: \emph{Proceedings of the AAAI Conference on Artificial
  Intelligence}, volume~36. pp. 2795--2803.
\newblock \doi{10.1609/aaai.v36i3.20183}.

\bibitem[{Xu et~al.(2022)Xu, Cai, He, Lin and Zhang}]{xu2022fastlio2}
Xu W, Cai Y, He D, Lin J and Zhang F (2022) {FAST-LIO2}: Fast direct
  {LiDAR}-inertial odometry.
\newblock \emph{IEEE Transactions on Robotics} 38(4): 2053--2073.
\newblock \doi{10.1109/TRO.2022.3141876}.

\bibitem[{Yin et~al.(2022)Yin, Li, Li, Yu and Zou}]{yin2022m2dgr}
Yin J, Li A, Li T, Yu W and Zou D (2022) {M2DGR}: A multi-sensor and
  multi-scenario {SLAM} dataset for ground robots.
\newblock \emph{IEEE Robotics and Automation Letters} 7(2): 2266--2273.
\newblock \doi{10.1109/LRA.2021.3138527}.

\bibitem[{Yoon et~al.(2019)Yoon, Tang and Barfoot}]{yoon2019mapless}
Yoon D, Tang T and Barfoot T (2019) Mapless online detection of dynamic objects
  in {3D} lidar.
\newblock In: \emph{Conference on Computer and Robot Vision (CRV)}. pp.
  113--120.
\newblock \doi{10.1109/CRV.2019.00023}.

\bibitem[{Zhang et~al.(2016)Zhang, Kaess and Singh}]{zhang2016degeneracy}
Zhang J, Kaess M and Singh S (2016) On degeneracy of optimization-based state
  estimation problems.
\newblock In: \emph{IEEE International Conference on Robotics and Automation
  (ICRA)}. pp. 809--816.
\newblock \doi{10.1109/ICRA.2016.7487211}.

\bibitem[{Zhang et~al.(2021)Zhang, Camurri, Wisth and
  Fallon}]{zhang2021newercollege}
Zhang L, Camurri M, Wisth D and Fallon M (2021) Multi-camera {LiDAR} inertial
  extension to the {Newer College} dataset.
\newblock arXiv:2112.08854.

\bibitem[{Zhang et~al.(2023)Zhang, Helmberger, Fu, Wisth, Camurri, Scaramuzza
  and Fallon}]{zhang2023hiltioxford}
Zhang L, Helmberger M, Fu LFT, Wisth D, Camurri M, Scaramuzza D and Fallon M
  (2023) {Hilti-Oxford} dataset: A millimeter-accurate benchmark for
  simultaneous localization and mapping.
\newblock \emph{IEEE Robotics and Automation Letters} 8(1): 408--415.
\newblock \doi{10.1109/LRA.2022.3226077}.

\bibitem[{Zhang and Scaramuzza(2018)}]{zhang2018tutorial}
Zhang Z and Scaramuzza D (2018) A tutorial on quantitative trajectory
  evaluation for visual(-inertial) odometry.
\newblock In: \emph{IEEE/RSJ International Conference on Intelligent Robots and
  Systems (IROS)}. pp. 7244--7251.
\newblock \doi{10.1109/IROS.2018.8593941}.

\bibitem[{Zheng and Zhu(2024)}]{zheng2024trajlo}
Zheng X and Zhu J (2024) {Traj-LO}: In defense of {LiDAR}-only odometry using
  an effective continuous-time trajectory.
\newblock \emph{IEEE Robotics and Automation Letters} 9(2): 1961--1968.
\newblock \doi{10.1109/LRA.2024.3352360}.

\end{thebibliography}

\end{document}